\def\year{2022}\relax
\documentclass[letterpaper]{article} 
\usepackage{aaai22}  
\usepackage{times}  
\usepackage{helvet}  
\usepackage{courier}  
\usepackage[hyphens]{url}  
\usepackage{graphicx} 
\usepackage{multirow}
\usepackage{subfigure}
\usepackage{natbib}  
\usepackage{caption} 
\DeclareCaptionStyle{ruled}{labelfont=normalfont,labelsep=colon,strut=off} 
\usepackage{algorithm}
\usepackage{algorithmic}

\usepackage{newfloat}
\usepackage{listings}
\floatstyle{ruled}
\newfloat{listing}{tb}{lst}{}
\floatname{listing}{Listing}
\title{Bridging between Cognitive Processing Signals and Linguistic Features via a Unified Attentional Network}
\author{
    Yuqi Ren,    Deyi Xiong\thanks{~Corresponding author} \\
}
\affiliations{
    College of Intelligence and Computing, Tianjin University, Tianjin, China, 300350\\

    \{ryq20, dyxiong\}@tju.edu.cn
}

\usepackage{bibentry}

\begin{document}

\maketitle

\begin{abstract}
Cognitive processing signals can be used to improve natural language processing (NLP) tasks. However, it is not clear how these signals correlate with linguistic information. Bridging between human language processing and linguistic features has been widely studied in neurolinguistics, usually via single-variable controlled experiments with highly-controlled stimuli. Such methods not only compromises the authenticity of natural reading, but also are time-consuming and expensive. In this paper, we propose a data-driven method to investigate the relationship between cognitive processing signals and linguistic features. Specifically, we present a unified attentional framework that is composed of embedding, attention, encoding and predicting layers to selectively map cognitive processing signals to linguistic features. We define the mapping procedure as a bridging task and develop 12 bridging tasks for lexical, syntactic and semantic features. The proposed framework only requires cognitive processing signals recorded under natural reading as inputs, and can be used to detect a wide range of linguistic features with a single cognitive dataset. Observations from experiment results resonate with previous neuroscience findings. In addition to this, our experiments also reveal a number of interesting findings, such as the correlation between contextual eye-tracking features and tense of sentence.
\end{abstract}

\section{Introduction}
Cognitively-inspired NLP which uses cognitive processing signals to enhance NLP models in a wide range of tasks, such as sentiment analysis \citep{barrett2018sequence}, dependency parsing \citep{strzyz2019towards}, named entity recognition \citep{DBLP:conf/naacl/HollensteinZ19}, part of speech tagging \citep{barrett2016weakly}, etc. From a computational perspective, cognitive processing signals can introduce additional information that underlies the ways that human brains comprehend texts \citep{DBLP:conf/ijcai/MathiasKMB20, DBLP:conf/aaai/MishraKB16, DBLP:conf/acl/RenX20}. Such information could be exploited to teach machines to process texts. However, how these cognitive processing signals resonate with linguistic information that is usually used by NLP models is not clear yet. Bridging between cognitive processing signals and linguistic features is certainly desirable for cognitively-inspired NLP.

In neurolinguistics, since the early discovery that the brain is directly involved in sentence comprehension in patients with brain damage, the process of language comprehension in the brain has been studied for hundreds of years \cite{1996Neuroscience}. Current studies have investigated the relationship between brain activities and a number of linguistic features, including basic features (e.g., Word Length, Word Frequency) \cite{rayner1998eye} and complex features (e.g., Semantic Surprisal, Syntactic or Semantic Ambiguities) \cite{rogalsky2009selective}. Most studies have conducted specially designed controlled experiments for specific linguistic features, and draw conclusions by analyzing differences in brain activities \cite{constable2004sentence, friederici2011brain}. For instance, in order to locate the region of the brain with a specific function of syntactic processing in sentence comprehension, researchers have compared and analyzed fMRI images that are recorded when subjects read sentences vs. meaningless word lists \cite{friederici2011brain}. In such controlled experiments, reading materials for testing are usually unnatural, lacking lexical and syntactic richness exhibited in real-world texts. Hence, the recorded cognitive processing signals are also unnatural. Moreover, the study of each type of linguistic features requires multiple records of cognitive processing signals, which is very labor-intensive and time-consuming. Thus, a unified and effective method to bridge between cognitive processing signals and linguistic features is also beneficial to neurolinguistics.

In this paper, we propose an efficient data-driven framework based on attention mechanism to study the relationship between linguistic features and cognitive processing signals. Different from neurolinguistic methods, the cognitive processing signals used in our experiments are all from natural reading materials, rather than highly-controlled stimuli. The results of our model are hence closer to reflect the actual text comprehension process. In order to bridge between cognitive processing signals and linguistic features, we propose an attentional framework to learn the importance of cognitive processing signals to linguistic features by feeding signals as inputs to the model to predict linguistic features. Compared with traditional feature selection methods (e.g., Mutual Information, Random Forests), the proposed attention mechanism can capture dependency between words. To model the relationship between cognitive processing signals and linguistic features, we adapt standard word-level attention into feature-level attention.

For a broad coverage of linguistic features and cognitive processing signals, we design a total of 12 bridging tasks that predict linguistic features from cognitive processing signal inputs. This basic methodology behind our model, which feeds representations into a classifier for linguistic prediction, has been widely used to study the interpretability of pre-training models in NLP \cite{DBLP:conf/acl/BaroniBLKC18, DBLP:conf/naacl/HewittM19}. According to the nature of linguistic features, we divide the bridging tasks into three categories: lexical, syntactic and semantic. For the generality of our bridging tasks, linguistic features used in our experiments can be easily obtained via off-the-shelf NLP tools. Yet another advantage of our model is that we can study the connections of cognitive processing signals to arbitrary linguistic features with only one dataset annotated with cognitive processing signals, without requiring to develop a large number of controlled experiments.

In a nutshell, our main contributions include:
\begin{itemize}
\item We propose a unified attentional framework to bridge between cognitive processing signals and linguistic features, which adapts traditional word-level attention to feature-level attention.
\item Twelve bridging tasks are developed to investigate the weighted alignments of eye-tracking and EEG signals to a wide range of linguistic features, including lexical, syntactic and semantic features.
\item Experiments provide new evidences for previous neurolinguistic findings. Additionally, our experiments exhibit new interesting findings, which are not present previously, on the underlying relations between cognitive processing signals and linguistic features. We demonstrate the effectiveness of our proposed method by conducting a number of comparison experiments from the computational perspective.
\end{itemize}

\section{Related Work}
\subsection{Relations between Linguistic Features and Cognitive Processing Signals}
In neurolinguistics, experiments have proven that cognitive processing signals can effectively reflect important textual information in language comprehension \cite{kilianska2021eye, van2010language}. \citet{rayner2004effect} have studied word recognition with eye-tracking signals. They ask participants to read highly-controlled stimuli, such as:
(1)  \emph{used a knife to chop the large carrots}. (2) \emph{used a pump to inflate the large carrots}. The target word is \emph{carrots} in all sentences. Obviously, the combination of verb and instrument is anomalous in sentence (2). The eye-tracking results show that the gaze duration of the target word in sentence (2) is significantly longer than that in sentence (1). It suggests that gaze duration is sensitive to semantic anomaly. Other methods that probe linguistic features based on Electroencephalogram (EEG) or functional magnetic resonance imaging (fMRI) signals in brain are similar to this research. These neurolinguistic studies usually develop variable-controlled experiments that only change the parameters of interest while other variables are kept intact. Hence, any changes in cognitive processing signals can be attributed to the variation \cite{weiss2003contribution, constable2004sentence}. \citet{mitchell2008predicting} map a word into a semantic feature space and analyze the distribution of each semantic feature in the brain by predicting the corresponding fMRI image of the word.

\subsection{Feature Selection}
The way that we aggregate information from cognitive signals using the attention mechanism and feed cognitive information to a classifier is related to feature selection. The general practice of feature selection is assigning an ``importance'' score to each feature. Based on the estimated scores, we can improve the performance of model by enhancing relatively important features or provide explanations for black-box model. This technique has been widely used in many different areas \cite{DBLP:books/kl/LiuM98, DBLP:journals/tkde/LiuY05}. The feature selection methods for classification can be roughly grouped into three categories: filtering methods, wrapper methods, and embedded methods \cite{DBLP:journals/bioinformatics/SaeysIL07}.

The filtering methods usually obtain feature importance scores by calculating the correlation between features and target tags, such as using mutual information \cite{DBLP:journals/ijprai/LiuYLXS21}, pearson correlation coefficient \cite{DBLP:conf/ciarp/CoelhoBV10}. These approaches have excellent scalability, but ignore the dependency between features. The wrapper methods generate and evaluate various feature subsets by training and testing a specific classification model, such as recursive feature elimination \cite{DBLP:conf/iih-msp/ZengCTA09}. Although these methods consider the dependency between features, they are usually computationally intensive. In the embedded methods, the process of feature scoring is built into classifiers, such as random forest \cite{DBLP:journals/bmcbi/Diaz-UriarteA06}, SVM \cite{DBLP:journals/iet-ipr/KhanSJAYS18}. Advantages of these methods include the interaction between features and classifiers as well as less computational cost than the wrapper methods. The attention mechanism used in this paper can be grouped into the embedded methods, which learn the degree of importance of features during neural model training.

\begin{figure*}[htb]
\centering
\includegraphics[width=0.65\textwidth, height=3.6cm]{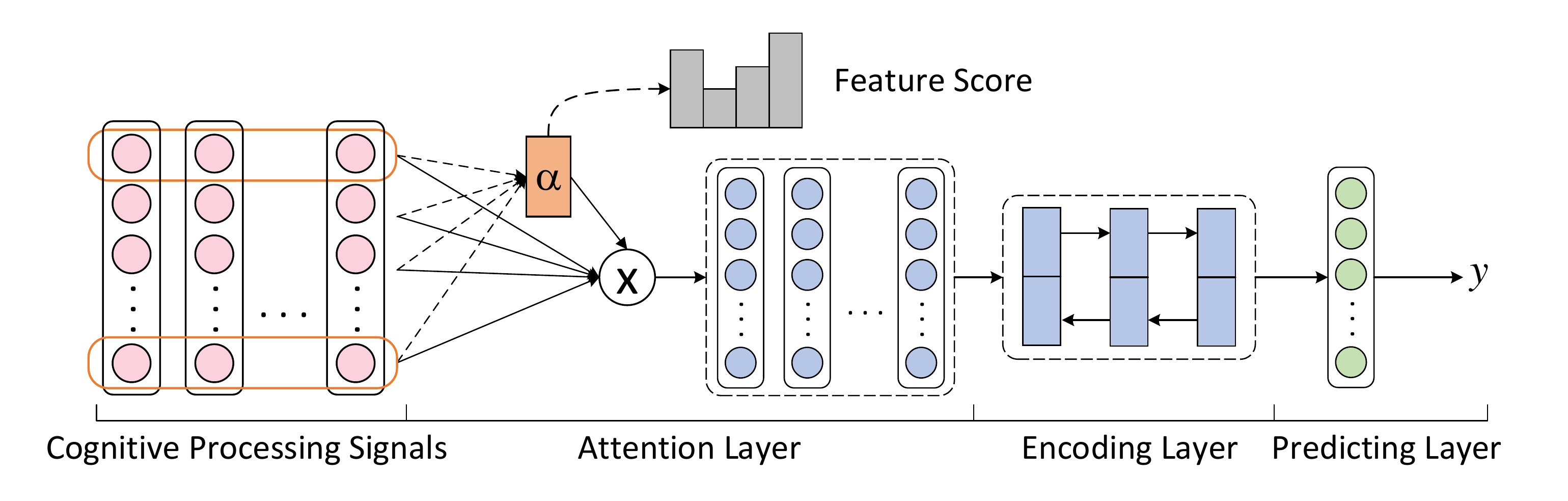}
\caption{Neural framework of the proposed attentional network for cognitive-linguistic bridging tasks. $\alpha$ represents the degree of importance of cognitive processing signals to the final prediction. ${y}$ is the final prediction of the model.}\label{model}
\end{figure*}

\section{The Unified Attentional Bridging Network}
Our model is an unified framework for bridging between different cognitive processing signals and a wide range of linguistic features. The bridging task is defined as a classification task that predicts linguistic features from cognitive processing signals. Since the attention mechanism has outstanding feature selection ability, we utilize the attention mechanism \cite{DBLP:conf/iclr/LinFSYXZB17} to capture the relationship between cognitive processing signals and linguistic features. The framework of our model is visualized in Figure \ref{model}, which is composed of four layers: input layer, attention layer, encoding layer and predicting layer.

\textbf{Cognitive Processing Signals.}
The inputs of our model are word-level cognitive processing signals that are recorded under a natural reading configuration. For a given sequence $X$, its distributed representation can be denoted as $H\in{R^{n\times{d}}}$, where $n$ and $d$ are the length of sentence and the dimension of cognitive processing signal embeddings, respectively. In this paper, we use two types of cognitive processing signals: eye-tracking and EEG. More details of these two types of signals will be presented in Experiments Section.

\textbf{Attention Layer.}
It is widely acknowledged that attention mechanism can capture semantic dependencies between words in sentences. In this paper, we aim to capture the sensitivity of each cognitive processing signal to linguistic features, so we change the traditional word-level attention into feature-level attention. The updated calculation process of attention is as follows:
\begin{equation}\label{equ0}
\alpha={\rm softmax}({\rm tanh}(W_{att}H+b_{att})v)
\end{equation}
where $W_{att}\in{R^{d\times{n}}}$, $b_{att}\in{R^{d}}$, $v\in{R^{d}}$ are trainable parameters in the model. $v$ is used to learn importance vectors where the dimension size is equal to the dimension size of cognitive processing signal embeddings. $\alpha{\in{R^{d}}}$ obtained after softmax normalization are real values that represent the degree of the importance of each cognitive processing signal to a target linguistic feature. The weighted sentence representation can be formulated as:
\begin{equation}\label{equ1}
H^{att}=H{\otimes}\alpha
\end{equation}
where $\otimes$ denotes the Hadamard product that multiplies elements at the same position.

\textbf{Encoding Layer.}
We stack a Bi-LSTM layer over the attentional layer to encode the new sentence representation to capture contextual information. The Bi-LSTM encoder is a concatenation of the forward and backward hidden states.

\textbf{Predicting Layer.}
If the target output is a sequence of labels, we use conditional random field (CRF) \cite{DBLP:conf/icml/LaffertyMP01} as the predictor. First, we map the output of Bi-LSTM $H^{'}$ into another semantic space, the dimension of which is equal to the number of output tags. For a given time step $i$, it can be formulated as follows:
\begin{equation}\label{equ2}
o_i=W_{s}H^{'}_i+b_{s}
\end{equation}
where $W_{s}$ and $b_{s}$ are trainable parameters. Then, we calculate the score of the entire predicted tag sequence $y$ as follows:
\begin{equation}\label{equ3}
score(y|X)=\sum^{n}_{i=1}(T_{i-1,i}+o_{i,y_i})
\end{equation}
where $T$ is a transition score matrix that refers to the transition probability of two successive labels. Finally, the model will output the tag sequence with the highest score estimated in Eq. \ref{equ3}.

For single-output classification tasks, we use a max-pooling to reduce the dimension of the output of Bi-LSTM and obtain a vector representation $h\in{R^d}$ for each sentence. The probability of a predicted label is obtained by a softmax function as follows:
\begin{equation}\label{equ4}
P(h)={\rm softmax}(W_{p}h+b_{p})
\end{equation}
where $W_{p}$ and $b_{p}$ are trainable parameters.
\begin{table*}[ht]
\footnotesize
\centering
\begin{tabular}{lll}
\hline
\multirow{2}{*}{\textbf{EARLY}}
& first fixation duration (FFD) & the duration of word $w$ that is first fixated\\
& first pass duration (FPD) & the sum of the fixations before eyes leave the word $w$\\
\hline
\multirow{6}{*}{\textbf{LATE}}
& number of fixations (NFIX) & the number of times word $w$ that is fixated \\
& fixation probability (FP) & the probability that word $w$ is fixated \\
& mean fixation duration (MFD) & the average fixation durations for word $w$\\
& total fixation duration (TFD) & the total duration of word $w$ that is fixated \\
& $n$ re-fixations (NR) & the number of times word $w$ that is fixated after the first fixation \\
& re-read probability (RRP) & the probability of word $w$ that is fixated more than once \\
\hline
\multirow{9}{*}{\textbf{CONTEXT}}
& total regression-from duration (TRD) & the total duration of regressions from word $w$ \\
& $w$-2 fixation probability ($w$-2 FP) & the fixation probability of the word $w$-2 \\
& $w$-1 fixation probability ($w$-1 FP) & the fixation probability of the word $w$-1 \\
& $w$+1 fixation probability ($w$+1 FP) & the fixation probability of the word $w$+1 \\
& $w$+2 fixation probability ($w$+2 FP) & the fixation probability of the word $w$+2 \\
& $w$-2 fixation duration ($w$-2 FD) & the fixation duration of the word $w$-2 \\
& $w$-1 fixation duration ($w$-1 FD) & the fixation duration of the word $w$-1 \\
& $w$+1 fixation duration ($w$+1 FD) & the fixation duration of the word $w$+1 \\
& $w$+2 fixation duration ($w$+2 FD) & the fixation duration of the word $w$+2 \\
\hline
\end{tabular}
\caption{Eye-tracking features used in this work.}\label{Tab:eye-tracking features}
\end{table*}
\section{Linguistic Features and Bridging Tasks}
To select specific linguistic features for our bridging tasks, we mainly follow two criteria: (1) \textbf{Interpretability}. The selected linguistic features should be interpretable and independent of each other. (2) \textbf{Extensibility}. The selected linguistic features can be automatically generated by off-the-shelf NLP tools so that these features can be easily annotated to the training data and bridging tasks can be easily extended to available cognitive datasets. In this way, we construct four bridging tasks for each group of linguistic features (i.e., lexical, syntactic, semantic).

\textbf{Lexical Features.}
Normally, the occurrence of complex words will disturb the comprehension of text. The first task is hence to predict the length of a word (\textbf{WordLen}). We use the average length of all words in a sentence as the prediction label. The second task in this group is to predict lexical density (\textbf{LD}) that refers to the ratio of the number of content words (e.g., Noun, Verb) to the total number of words in a sentence. This task allows us to analyze the response of cognitive processing signals to content words and function words. The third bridging task is for predicting degree of polysemy (\textbf{DP}), which is the total number of senses contained by each word in the sentence. Intuitively, the more meanings of words in a sentence, the harder the comprehension of the text is. The senses of word can be automatically induced from WordNet\footnote{https://wordnet.princeton.edu/.} . The final out-of-vocabulary (\textbf{OOV}) task in this group is to sum the number of words in a sentence, which do not occur in the list of common words: the combination of the General Word Service List\footnote{http://jbauman.com/gsl.html.} and the Academic WordList\footnote{http://www.victoria.ac.nz/lals/resources/academicwordlist/.} . The purpose of this task is to investigate what remarkable cognitive processing variations might appear when unfamiliar words are present. Since the WordLen and LD tasks are correlated with sentence length, we alleviate the noise via normalization according to sentence length. To avoid the problem of data imbalance, the outputs of each task are three discrete labels from three different intervals segmented according to the values of these features (e.g., DP, OOV). Each label has the equal number of training instances. We regard these tasks as a three-class classification problem.

\textbf{Syntactic Features.}
Syntactic features focus on how varied and sophisticated sentence elements and their structures are \cite{lu2010automatic}. The first task in this group is to predict complex nominals per clause (\textbf{CNC}), which is defined as the ratio of the number of complex nominals to the number of clauses. Complex nominals, e.g., nominal groups or nominal clauses, are frequently used to measure the complexity of a sentence in English texts \cite{nakov2013interpretation}. We use L2 Syntactic Complexity Analyzer tool\footnote{http://www.personal.psu.edu/xxl13/downloads/l2sca.html.} to obtain the annotation of this feature. The second task is predicting the length of a sentence (\textbf{SenLen}), aiming to investigate the sensitivity of different cognitive processing signals to sentence length. The above two tasks are also recast as a three-class classification problem. The third task is predicting part-of-speech (\textbf{POS}) tags, designed to check the differences in human cognition processing of different parts of speech in sentences. POS tags are obtained by using Stanford Parser\footnote{https://nlp.stanford.edu/software/lex-parser.shtml.} . The bigram shift (\textbf{BShift}) task tests the extent to which different cognitive processing signals are sensitive to grammatical errors related to word order. This task is a binary classification problem, where we randomly swap two adjacent words in a sentence as a negative sample while the original sentence is regarded as a positive sample.

\textbf{Semantic Features.}
We define semantic features of a sentence as features containing meaningful information obtained from context. In this paper, semantic features for bridging tasks are mainly inspired by \citet{DBLP:conf/acl/BaroniBLKC18}. The first task in this group is to detect the tense of a sentence, which is usually dependent on the part-of-speech of the verb in the main clause: VBP/VBZ/VBG labeled as the present tense, VBD/VBN as the past tense, and VBC/VBF as the future tense. The subject number (\textbf{SubjNum}) task and object number (\textbf{ObjNum}) task focus on the number of subjects and objects of a sentence, respectively. These two tasks are to identify whether the target is singular or plural. The Tense, SubjNum and ObjNum are considered as semantic tasks rather than syntactic tasks, since these tasks essentially require understanding the meaning of a given sentence (e.g., whether the event described in the sentence occurred in the past). The changes of the labels of these tasks do not result in any changes in sentence structures. Discourse Connector Count (\textbf{DCC}) task is to predict the number of discourse connectors (e.g., \emph{however}, \emph{because}) that logically connects different discourse units. These connectors usually have causal or inferential implications. We aim to explore the sensitivity of cognitive processing signals to discourse coherence by this task. The Discourse Connector List\footnote{https://www.eapfoundation.com/vocab/academic/other/dcl/.} is used to annotate this task. Similar to the previous lexical tasks, DCC is also set as a three-class classification task.



\begin{table*}[ht]
\footnotesize
\centering
\begin{tabular}{llcccccccccccc}
\hline
\multirow{2}{*}{\textbf{Type}} &
\multirow{2}{*}{\textbf{CF}} &
\multicolumn{12}{c}{\textbf{Bridging Tasks}} \\
~ & ~ & \scriptsize LD & \scriptsize WordLen & \scriptsize DP & \scriptsize OOV & \scriptsize CNC & \scriptsize SenLen & \scriptsize POS & \scriptsize Bshift & \scriptsize Tense & \scriptsize SubjNum & \scriptsize ObjNum & \scriptsize DCC \\
\hline
\multirow{17}{*}{eye}
&FFD & 0.056 & 0.158 & 0.130 & 0.161 & 0.037 & 0.021 & 0.049 & 0.026 & 0.029 & 0.035 & 0.069 & 0.040 \\
&FPD & 0.068 & \textbf{0.226} & 0.121 & 0.076 & 0.031 & 0.054 & 0.104 & 0.047 & 0.034 & 0.057 & 0.088 & 0.016 \\
&NFIX & 0.070 & 0.070 & 0.005 & 0.085 & \textbf{0.152} & 0.072 & 0.032 & 0.055 & 0.029 & 0.067 & 0.041 & \textbf{0.120} \\
&FP & 0.034 & 0.050 & 0.013 & 0.030 & 0.044 & 0.035 & 0.042 & 0.039 & 0.028 & 0.080 & 0.080 & 0.066 \\
&MFD & 0.052 & 0.003 & 0.013 & 0.030 & 0.028 & 0.023 & \textbf{0.142} & 0.129 & 0.032 & \textbf{0.106} & 0.060 & 0.020 \\
&TFD & 0.093 & 0.032 & \textbf{0.206} & 0.180 & 0.123 & 0.028 & 0.060 & 0.060 & 0.026 & 0.075 & \textbf{0.109} & 0.016 \\
&NR & 0.076 & 0.094 & 0.182 & \textbf{0.223} & 0.124 & 0.060 & 0.047 & 0.041 & 0.029 & 0.054 & 0.046 & 0.015 \\
&RRP & 0.090 & 0.008 & 0.003 & 0.126 & 0.034 & 0.011 & 0.044 & 0.021 & 0.029 & 0.048 & 0.048 & 0.086 \\
&TRD & 0.001 & 0.052 & 0.002 & 0.013 & 0.041 & 0.009 & 0.032 & 0.049 & 0.028 & 0.062 & 0.033 & 0.014\\
&$w$-2 FP & 0.002 & 0.049 & 0.006 & 0.000 & 0.023 & \textbf{0.146} & 0.078 & 0.050 & 0.078 & 0.054 & 0.050 & 0.026 \\
&$w$-1 FP & 0.026 & 0.066 & 0.102 & 0.000 & 0.068 & 0.144 & 0.056 & 0.093 & 0.126 & 0.064 & 0.070 & 0.036 \\
&$w$+1 FP & 0.126 & 0.017 & 0.112 & 0.005 & 0.089 & 0.101 & 0.075 & \textbf{0.137} & 0.078 & 0.032 & 0.061 & 0.066 \\
&$w$+2 FP & \textbf{0.127} & 0.009 & 0.002 & 0.011 & 0.055 & 0.130 & 0.057 & 0.051 & 0.078 & 0.049 & 0.062 & 0.086 \\
&$w$-2 FD & 0.002 & 0.057 & 0.004 & 0.013 & 0.031 & 0.011 & 0.045 & 0.034 & \textbf{0.132} & 0.038 & 0.030 & 0.109 \\
&$w$-1 FD & 0.093 & 0.031 & 0.003 & 0.017 & 0.017 & 0.032 & 0.058 & 0.094 & 0.064 & 0.050 & 0.050 & 0.078 \\
&$w$+1 FD & 0.082 & 0.074 & 0.002 & 0.005 & 0.075 & 0.069 & 0.029 & 0.045 & 0.099 & 0.049 & 0.062 & 0.106 \\
&$w$+2 FD & 0.003 & 0.004 & 0.093 & 0.013 & 0.028 & 0.053 & 0.050 & 0.031 & 0.082 & 0.081 & 0.040 & 0.098 \\
\hline
\multirow{8}{*}{EEG}
&t1 & \textbf{0.253} & 0.134 & \textbf{0.237} & 0.159 & 0.106 & 0.046 & 0.125 & 0.123 & 0.074 & 0.116 & 0.123 & 0.101 \\
&t2 & 0.178 & 0.128 & 0.228 & 0.198 & 0.159 & 0.053 & 0.115 & 0.087 & 0.094 & 0.124 & 0.096 & 0.088 \\
&a1 & 0.093 & \textbf{0.200} & 0.023 & 0.082 & 0.130 & 0.132 & 0.180 & 0.125 & 0.145 & 0.133 & \textbf{0.175} & 0.120 \\
&a2 & 0.103 & 0.143 & 0.039 & \textbf{0.201} & 0.113 & 0.124 & 0.113 & 0.096 & 0.150 & \textbf{0.182} & 0.132 & 0.106 \\
&b1 & 0.068 & 0.037 & 0.116 & 0.089 & \textbf{0.167} & 0.133 & 0.091 & 0.116 & 0.177 & 0.086 & 0.106 & 0.140 \\
&b2 & 0.177 & 0.053 & 0.130 & 0.046 & 0.150 & 0.152 & 0.131 & 0.099 & \textbf{0.207} & 0.126 & 0.145 & 0.089 \\
&g1 & 0.102 & 0.154 & 0.145 & 0.099 & 0.075 & \textbf{0.187} & 0.158 & 0.147 & 0.086 & 0.125 & 0.122 & 0.153 \\
&g2 & 0.025 & 0.144 & 0.081 & 0.125 & 0.100 & 0.173 & 0.088 & \textbf{0.207} & 0.067 & 0.107 & 0.100 & \textbf{0.202} \\
\hline
\end{tabular}
\caption{Results of bridging between linguistic features (i.e., lexical, syntactic or semantic) and eye-tracking \& EEG signals. `CF': Cognitive Feature.}\label{Tab:attention_score}
\end{table*}

\section{Experiments}
\subsection{Dataset and Cognitive Processing Signals}
We used a cognitive dataset curated under natural reading: Zurich Cognitive Language Processing Corpus (ZuCo) \cite{hollenstein2018zuco}. It is a publicly available dataset\footnote{https://osf.io/q3zws/} of simultaneous eye-tracking and EEG signals from subjects reading natural sentences. This corpus includes recordings of 12 adult and native speakers reading approximately 1100 English sentences.

The dataset includes two reading paradigms: normal reading and task-specific reading. In the latter, subjects are asked to exercise some comprehension task before recording. The goal of this work is to analyze real language comprehension in natural reading. Therefore, we only used the data of normal reading. The reading materials in this paradigm consist of 700 sentences.

\textbf{Eye-tracking.}
The eye-tracking data of ZuCo are collected by an infrared video-based eye tracker EyeLink 1000 Plus. Since we want to cover all eye movement features, in addition to the measured fixation duration features, we also added the fixation probability which plays an important role in syntactic understanding \cite{demberg2008data}. Moreover, the fixation features of precursors and postcursors (e.g., w-1 fixation duration) are considered in this work. In total, we used 17 eye-tracking features that cover all stages of gaze behaviors to probe the linguistic features in eye-tracking signals. These features are divided into three groups based on reading stage: \textbf{EARLY}, the gaze behaviors when a word is first fixated, which reflect early word acquisition and syntactic processing; \textbf{LATE}, the gaze behaviors over a word after sentence reading, which generally reflect late syntactic processing and semantic understanding of sentences; \textbf{CONTEXT}, the eye-tracking features over neighboring words of the current word. More details of these 17 features are shown in Table \ref{Tab:eye-tracking features}.

\textbf{EEG.}
EEG signals measure voltage fluctuations in the cerebral cortex with high temporal resolution. The EEG signals in ZuCo are recorded by a 128-channel EEG Geodesic Hydrocel system. Each EEG record contains 128 electrode values where 23 EEG signals are removed since they are used to detect muscular artifacts \citep{hollenstein2018zuco}. Based on the frequency of brain's electrical signals, the left 105 EEG signals are divided into 8 frequency bands: $theta1$ (t1, 4-6 Hz), $theta2$ (t2, 6.5-8 Hz), $alpha1$ (a1, 8.5-10 Hz), $alpha2$ (a2, 10.5-13 Hz), $beta1$ (b1, 13.5-18 Hz), $beta2$ (b2, 18.5-30 Hz), $gamma1$ (g1, 30.5-40 Hz) and $gamma2$ (g2, 40-49.5 Hz). All EEG signals are normalized and averaged over all subjects. In this work, we used 8 EEG features that are obtained by averaging the 105 EEG signals at each frequency band to represent 8 frequency bands.

\begin{figure*}[ht]
\centering
\subfigure[Lexical tasks]
{
	\includegraphics[width=0.28\textwidth]{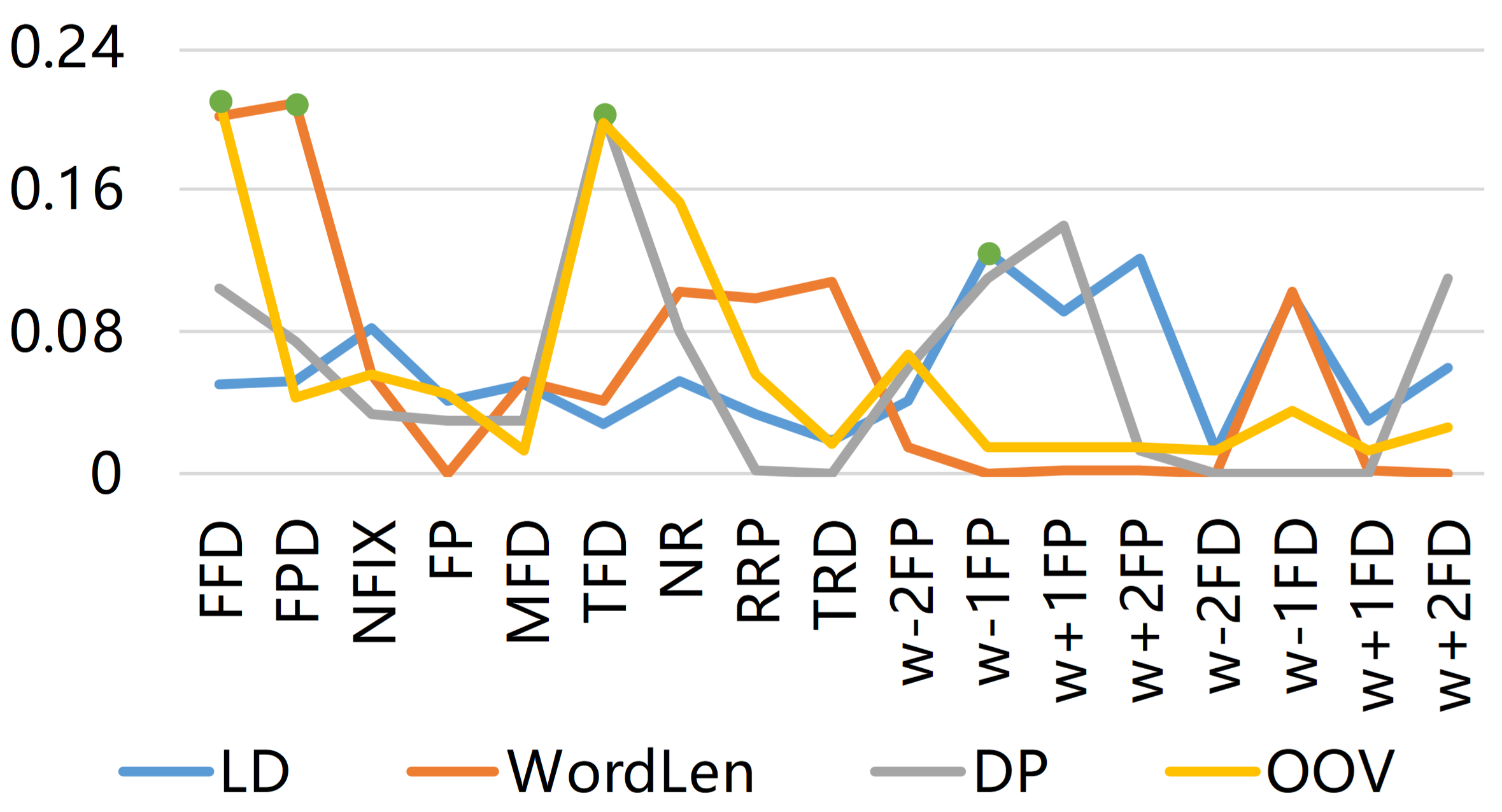}
}
\subfigure[Syntactic tasks]
{
	\includegraphics[width=0.28\textwidth]{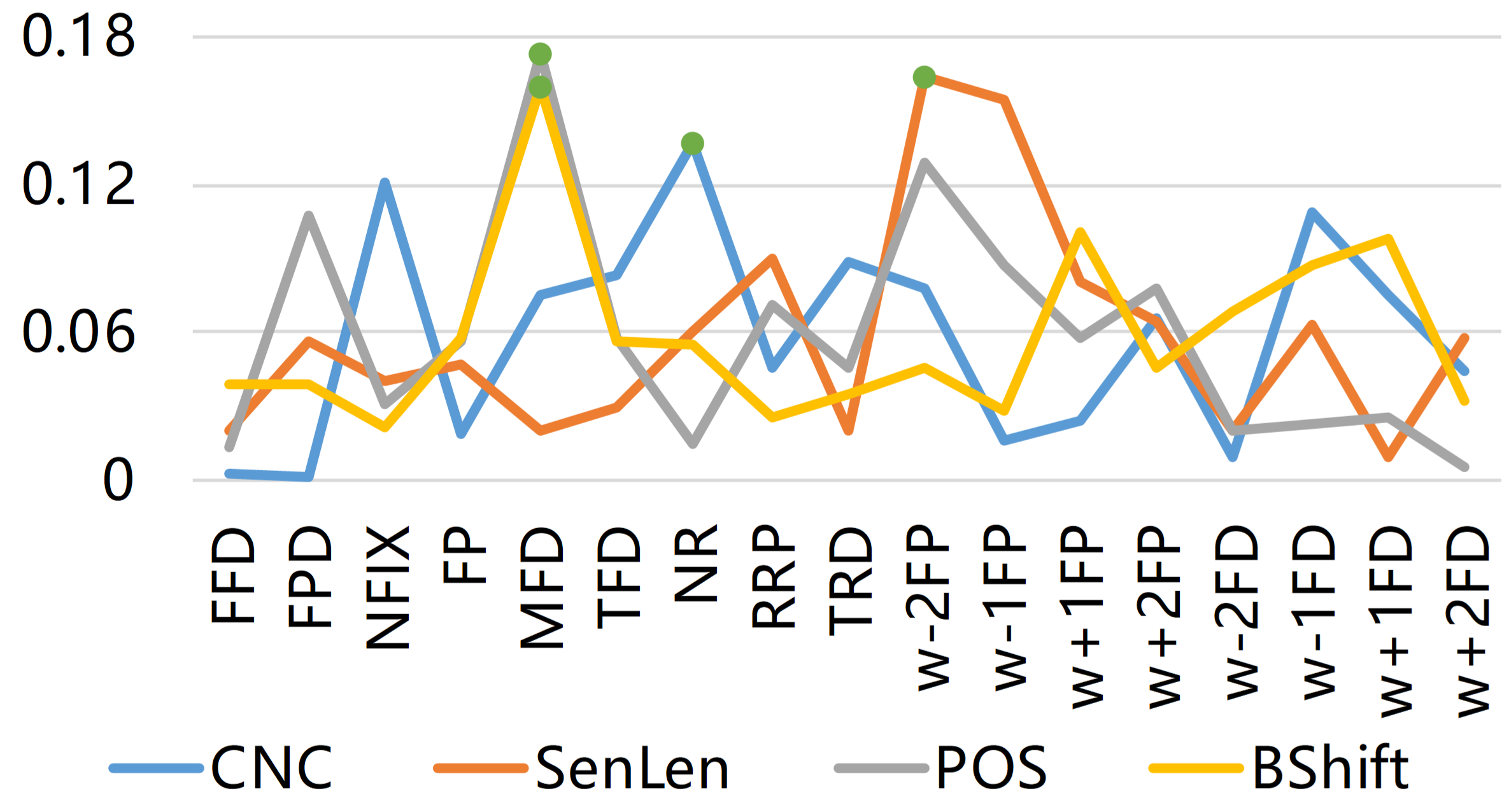}
}
\subfigure[Semantic tasks]
{
	\includegraphics[width=0.28\textwidth]{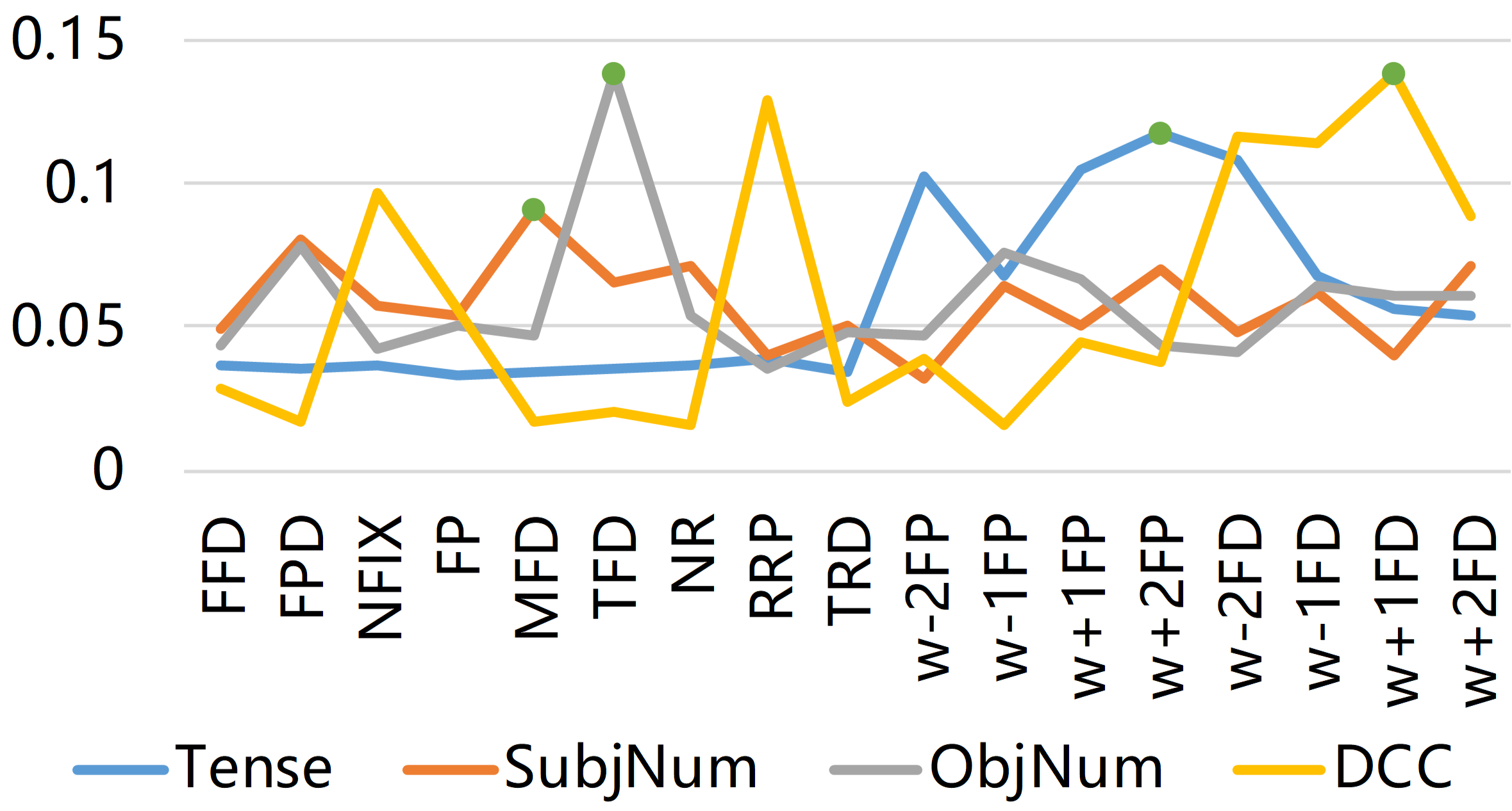}
}
\caption{Ablation results of bridging between linguistic features and eye-tracking signals with a model variant without the Bi-LSTM encoder. The x-coordinate denotes cognitive features while y-coordinate denotes attention scores after normalization. We highlight cognitive features with the highest attention scores for each bridging task with green dots.}\label{no Bi-LSTM}
\end{figure*}
\subsection{Experiment Settings}
The dimension of hidden states in the Bi-LSTM encoder was set to 20. Due to the data imbalance issue in Tense, SubjNum and ObjNum bridging tasks, we used focal loss \cite{DBLP:conf/iccv/LinGGHD17} to reduce the proportion of easily classified samples in the training. The cross-entropy loss function was used for other bridging tasks. To obtain robust experimental results, we performed 5-fold cross validation for all bridging tasks.
\begin{figure*}[ht]
\centering
\subfigure[Lexical tasks (Eye)]
{
	\includegraphics[width=0.28\textwidth]{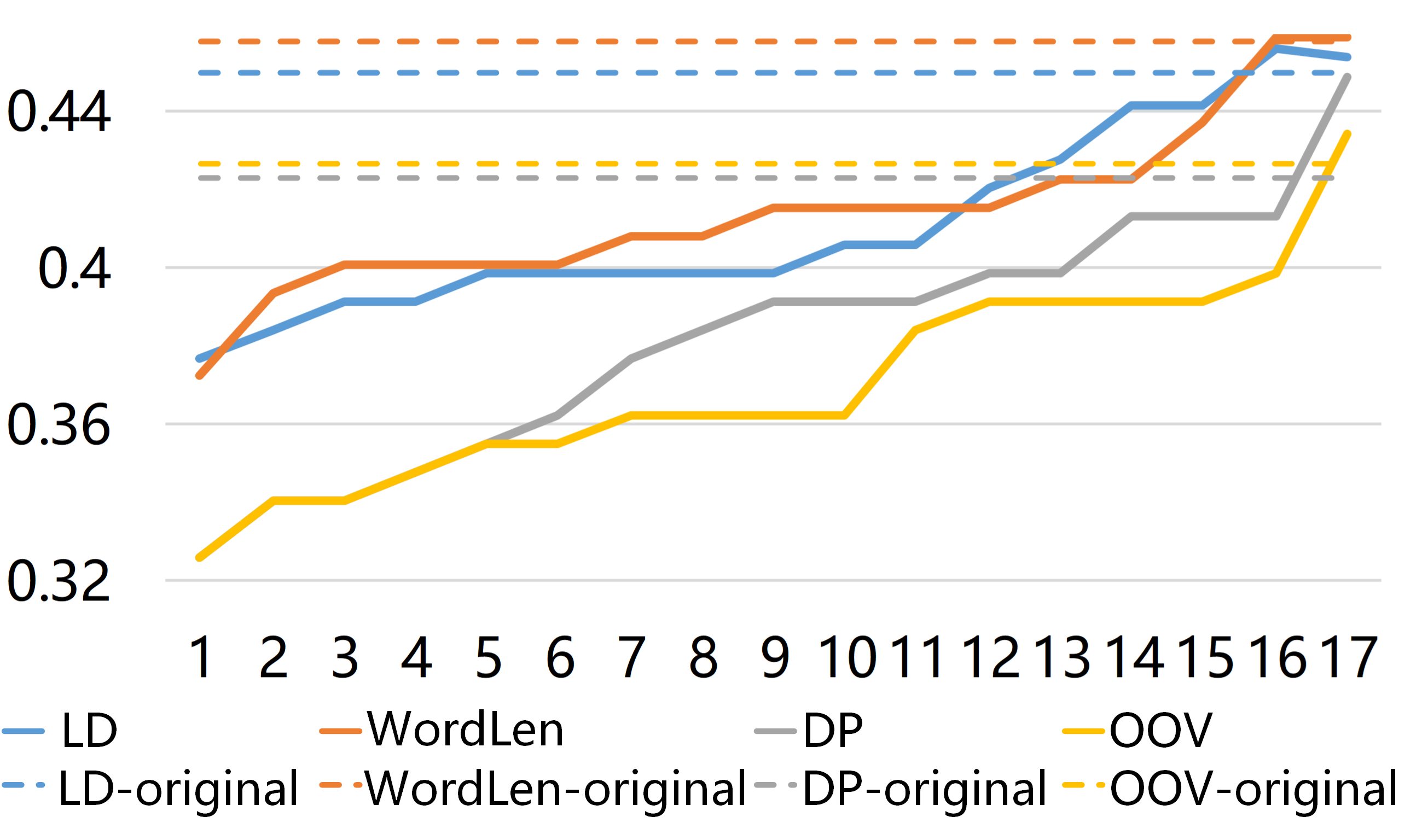}
}
\subfigure[Syntactic tasks (Eye)]
{
	\includegraphics[width=0.28\textwidth]{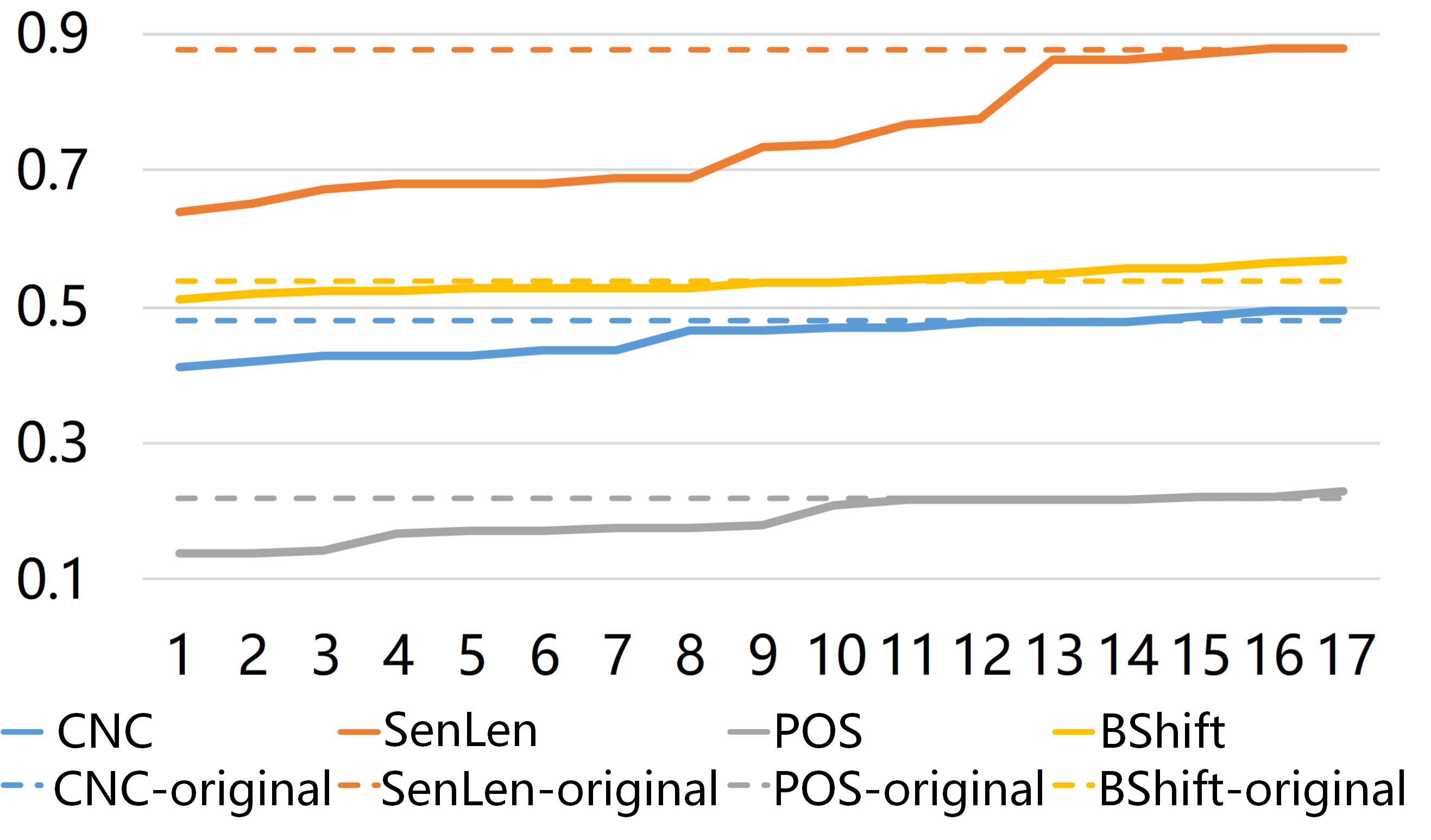}
}
\subfigure[Semantic tasks (Eye)]
{
	\includegraphics[width=0.28\textwidth]{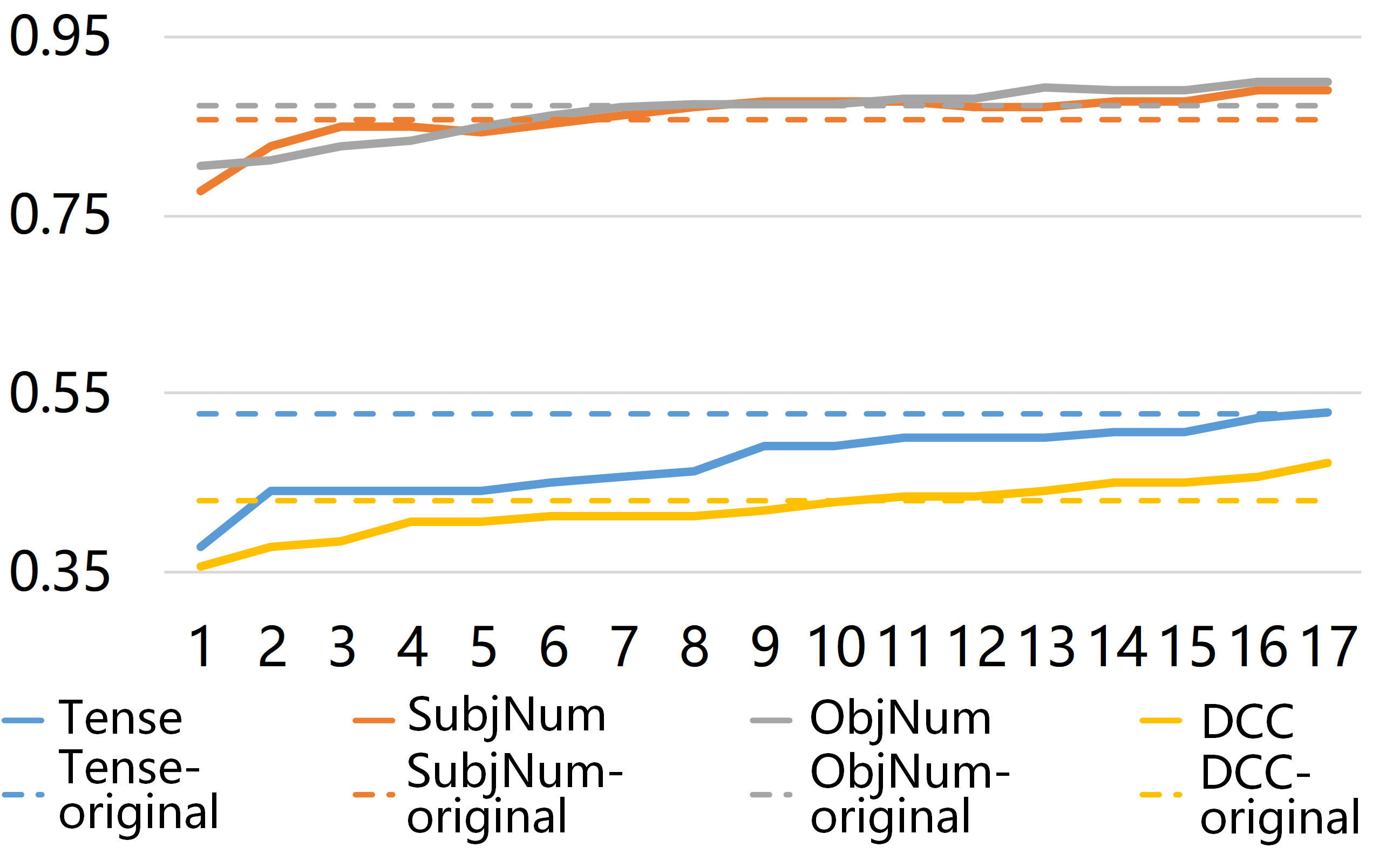}
}
\subfigure[Lexical tasks (EEG)]
{
	\includegraphics[width=0.28\textwidth]{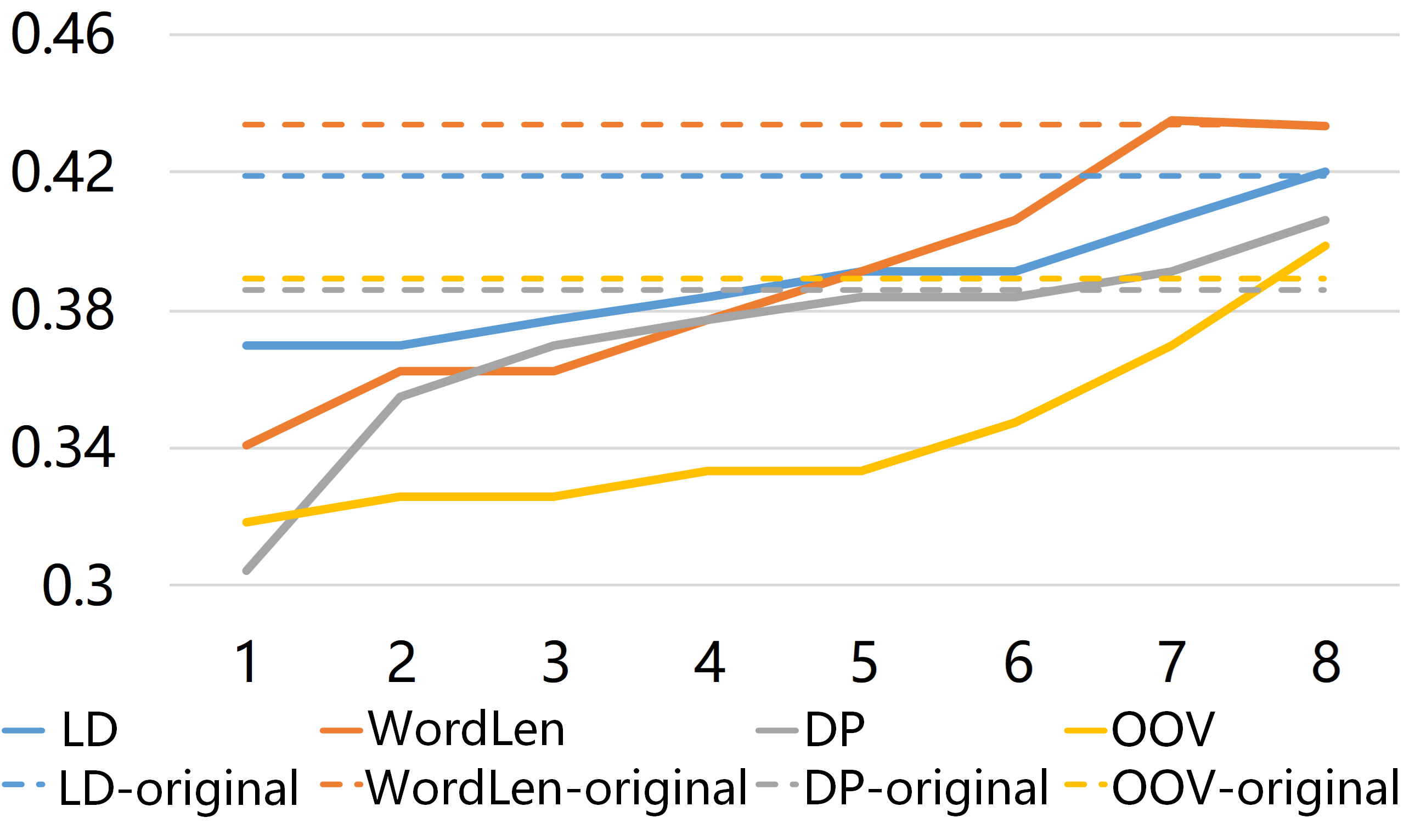}
}
\subfigure[Syntactic tasks (EEG)]
{
	\includegraphics[width=0.28\textwidth]{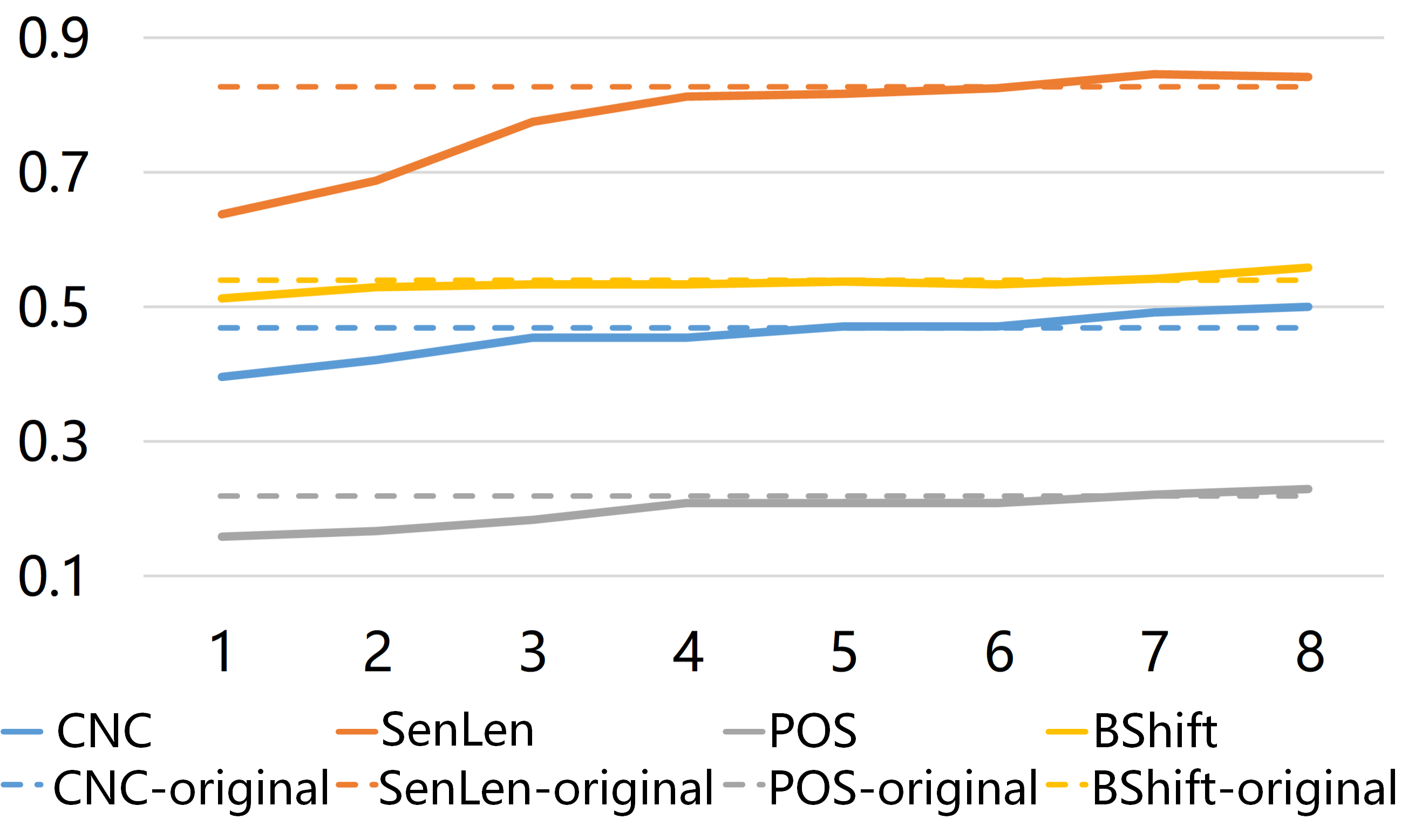}
}
\subfigure[Semantic tasks (EEG)]
{
	\includegraphics[width=0.28\textwidth]{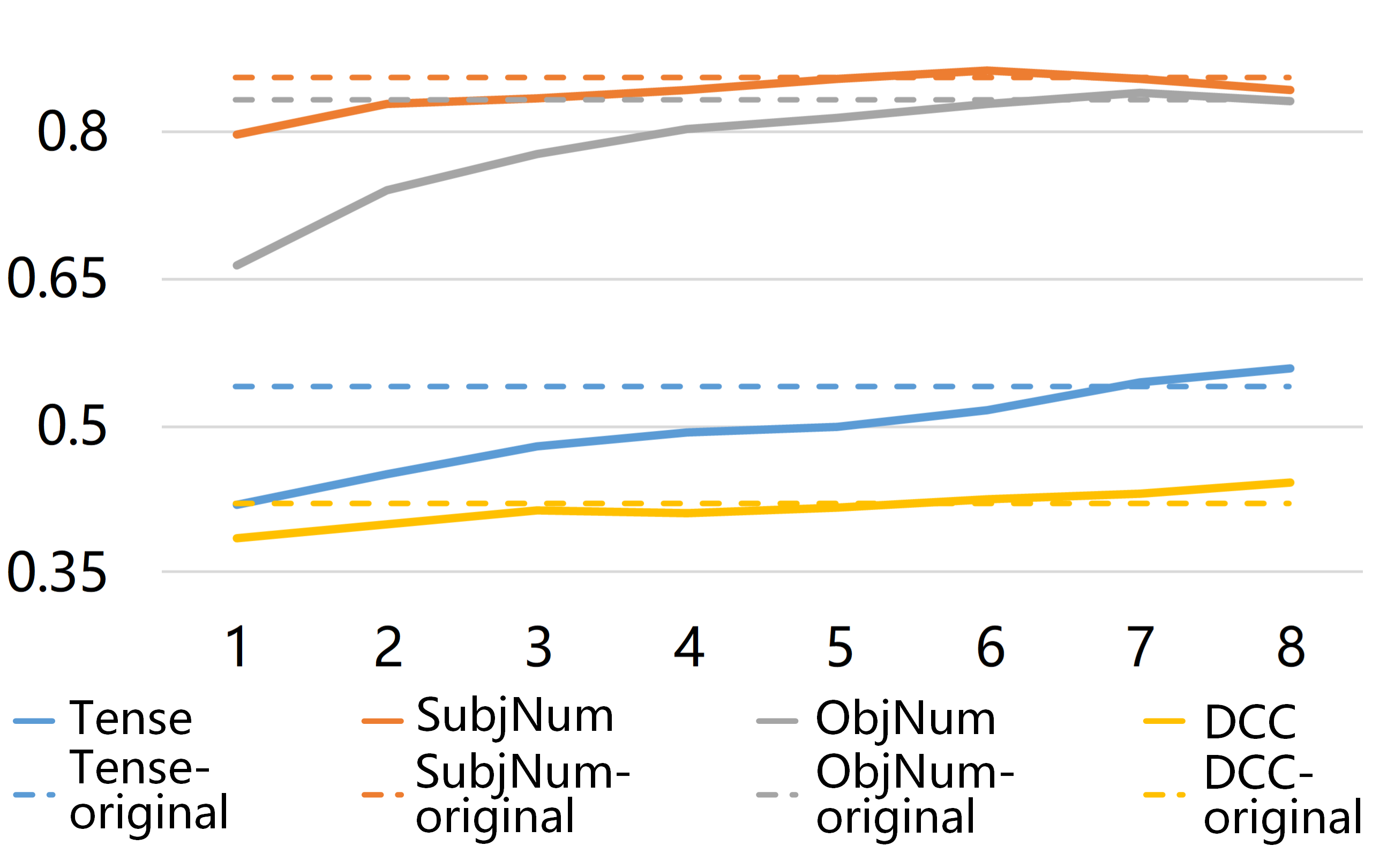}
}
\caption{Signal masking experiment results with eye-tracking and EEG signals. The x-coordinate denotes the attention score rankings of masked signals, and y-coordinate denotes the evaluation metrics of masking experiments ($F_1$ value). The dashed line represents the initial experimental results without masking (denoted as X-original).}\label{fig:2}
\end{figure*}

\begin{table*}[ht]
\small
\centering
\begin{tabular}{llcccccccccccc}
\hline
\multirow{3}{*}{\textbf{Type}} &
\multirow{3}{*}{\textbf{Cognitive Feature}} &
\multicolumn{12}{c}{\textbf{Feature Selection Method}} \\
~& ~ & \multicolumn{3}{c}{Mutual Information} & \multicolumn{3}{c}{RFE} & \multicolumn{3}{c}{Random Forest} & \multicolumn{3}{c}{Our Model} \\
\cline{3-14}
~ & ~ & LD & CNC & Tense & LD & CNC & Tense & LD & CNC & Tense & LD & CNC & Tense \\
\hline
\multirow{17}{*}{eye}
& FFD & 0.089 & 0.018 & 0.061 & 12 & 10 & 11 & 0.033 & 0.034 & 0.070 & 0.056 & 0.037 & 0.029 \\
& FPD & 0.044 & 0.017 & 0.056 & \textbf{1} & 11 & 14 & 0.069 & 0.056 & 0.041 & 0.068 & 0.031 & 0.034 \\
& NFIX & 0.069 & \textbf{0.180} & 0.034 & 9 & 8 & 2 & 0.071 & 0.065 & 0.036 & 0.070 & \textbf{0.152} & 0.029 \\
& FP & 0.159 & 0.146 & 0.039 & 15 & 3 & 4 & \textbf{0.136} & 0.084 & 0.081 & 0.034 & 0.044 & 0.028 \\
& MFD & 0.009 & 0.051 & 0.133 & 17 & 15 & 17 & 0.037 & 0.060 & 0.077 & 0.052 & 0.028 & 0.032 \\
& TFD & 0.065 & 0.015 & 0.026 & 11 & 7 & 6 & 0.050 & 0.047 & 0.026 & 0.093 & 0.123 & 0.026 \\
& NR & 0.034 & 0.089 & 0.020 & 7 & 13 & 15 & 0.045 & 0.055 & \textbf{0.111} & 0.076 & 0.124 & 0.029 \\
& RRP & 0.007 & 0.017 & 0.035 & 5 & 9 & 16 & 0.040 & \textbf{0.096} & 0.046 & 0.090 & 0.034 & 0.029 \\
& TRD & 0.007 & 0.156 & 0.040 & 6 & 5 & 10 & 0.035 & 0.039 & 0.039 & 0.001 & 0.041 & 0.028 \\
& $w$-2 FP & 0.012 & 0.021 & 0.055 & 16 & 16 & 12 & 0.048 & 0.050 & 0.050 & 0.002 & 0.023 & 0.078 \\
& $w$-1 FP & \textbf{0.168} & 0.022 & 0.028 & 13 & 14 & 9 & 0.068 & 0.056 & 0.068 & 0.026 & 0.068 & 0.126 \\
& $w$+1 FP & 0.075 & 0.051 & 0.032 & 14 & 4 & 8 & 0.074 & 0.057 & 0.074 & 0.126 & 0.089 & 0.078 \\
& $w$+2 FP & 0.007 & 0.032 & \textbf{0.144} & 8 & 6 & 5 & 0.062 & 0.057 & 0.100 & \textbf{0.127} & 0.055 & 0.078 \\
& $w$-2 FD & 0.039 & 0.015 & 0.117 & 10 & 12 & 13 & 0.040 & 0.070 & 0.029 & 0.002 & 0.031 & \textbf{0.132} \\
& $w$-1 FD & 0.061 & 0.039 & 0.056 & 2 & 17 & \textbf{1} & 0.063 & 0.064 & 0.051 & 0.093 & 0.017 & 0.064 \\
& $w$+1 FD & 0.088 & 0.063 & 0.080 & 3 & 2 & 7 & 0.062 & 0.063 & 0.042 & 0.082 & 0.075 & 0.099 \\
& $w$+2 FD & 0.066 & 0.065 & 0.044 & 4 & \textbf{1} & 3 & 0.066 & 0.048 & 0.058 & 0.003 & 0.028 & 0.082 \\
\hline
\multirow{8}{*}{EEG}
& t1 & 0.036 & 0.072 & 0.220 & 3 & \textbf{1} & 4 & 0.108 & 0.139 & 0.130 & \textbf{0.253} & 0.106 & 0.074 \\
& t2 & 0.229 & 0.068 & 0.118 & 6 & 7 & 7 & 0.128 & 0.134 & 0.114 & 0.178 & 0.159 & 0.094 \\
& a1 & 0.036 & 0.199 & 0.058 & 5 & 4 & \textbf{1} & 0.097 & 0.129 & 0.146 & 0.093 & 0.130 & 0.145 \\
& a2 & 0.135 & 0.068 & 0.165 & 7 & 8 & 3 & 0.113 & \textbf{0.144} & \textbf{0.159} & 0.103 & 0.113 & 0.150 \\
& b1 & 0.043 & 0.068 & 0.058 & 8 & 3 & 8 & 0.083 & 0.133 & 0.095 & 0.068 & \textbf{0.167} & 0.177 \\
& b2 & 0.115 & \textbf{0.390} & 0.088 & 4 & 4 & 5 & \textbf{0.166} & 0.116 & 0.114 & 0.177 & 0.150 & \textbf{0.207} \\
& g1 & 0.114 & 0.068 & 0.068 & \textbf{1} & 2 & 2 & 0.147 & 0.098 & 0.124 & 0.102 & 0.075 & 0.086 \\
& g2 & \textbf{0.294} & 0.068 & \textbf{0.225} & 2 & 6 & 6 & 0.158 & 0.106 & 0.119 & 0.025 & 0.100 & 0.067 \\
\hline
\end{tabular}
\caption{Feature importance scores estimated by our model and other feature selection methods on bridging tasks (LD, CNC and Tense) with eye-tracking and EEG signals. `RFE' denotes Recursive Feature Elimination. Since the Recursive Feature Elimination method selects features by recursively shrinking the sets of features, we only obtain the ranking of each feature among all features.}\label{Tab:ranking}
\end{table*}

\subsection{Results}

The experiment results are shown in Table \ref{Tab:attention_score}. From the results of eye-tracking signals, we have the following observations. First, FPD and FFD have relatively high attention scores on WordLen and OOV features, indicating that the perception of word length and word proficiency appears in the brain when the word if first fixated. Second, NR and TFD are highly correlated with DP and OOV. This observation is in accord with previous neurolinguistic finding that the difficulty of word understanding strongly influences the number of regressions to the word \cite{duffy1992eye}. Third, not surprisingly, the late eye movement behaviors (e.g., NFIX, TFD) are aligned to CNC feature that detects complex nominals. Fourth, it is worth noting that the probability of contextual words being fixated plays an important role in multiple bridging tasks (i.e., SenLen, Bshift, Tense, DCC). This is consistent with the finding that the constraints of contextual words affect the fixation duration and skipping frequency of the current word \cite{rayner1996effects}. We therefore suggest that eye movement behaviors over contextual words could be used to assess the syntactic and semantic complexity of sentences. Finally, we find that late eye movement behaviors (e.g., MFD, TFD) over words are more important for the judgment of subject and object singularity or plurality in sentences than early eye movement behaviors.

In summary, early eye movement behaviors focus on lexical features while eye movement behaviors over contextual words are sensitive to syntactic and semantic information. On the other hand, the eye-tracking features of late eye movement behaviors seem to correlate with all three types of linguistic features tested in our experiments, lexical, syntactic and semantic features.

EEG signals that record neuronal activity at a millisecond time scale can provide dynamic temporal information for brain language processing. Studying the oscillatory changes in electrical signals in the brain can help us understand various reading comprehension processes. Results of EEG are also consistent with previous neuroscience findings. (1) As Table \ref{Tab:attention_score} shows, compared with other frequency range, the attention scores of t1 and t2 are generally higher than other EEG features in the lexical bridging tasks, especially in LD task. This conforms with early insights that both open-class (OC) words (e.g., nouns, verbs) and closed-class (CC) words (e.g., articles, determiners) elicit a power increase in the theta frequency range, but OC words (related to the LD feature) have stronger power changes than CC words \cite{bastiaansen2005theta}. (2) The $theta$ and $beta$ frequency range are good at CNC bridging task, resonating with the previous finding that there is a large gap in $beta$ frequency range between sentences with complex syntax and sentences with simple syntax  \cite{weiss2005increased}. (3) The high frequency band ($gamma$) is acknowledged to be related to semantic information \cite{weiss2003contribution}. The results on DCC, which demonstrate the significance of $gamma$ frequency band in DCC bridging task, clearly provide yet another evidence for this.

More interestingly, the $gamma$ frequency band is sensitive to word order (BShift task), and $alpha$ frequency band is related to the number of subjects and objects in sentences.

Overall, these results on eye-tracking and EEG signals provide new evidences for previous neuroscience findings from a data-driven and computational perspective. These results also suggest that cognitive processing signals are related to rich linguistic information, which can be explored in various NLP tasks. Our model provides a unified and efficient way to probe linguistic features behind cognitive processing signals.

\begin{figure*}[ht]
\centering
\subfigure[LD task (Eye)]
{
	\includegraphics[width=0.28\textwidth]{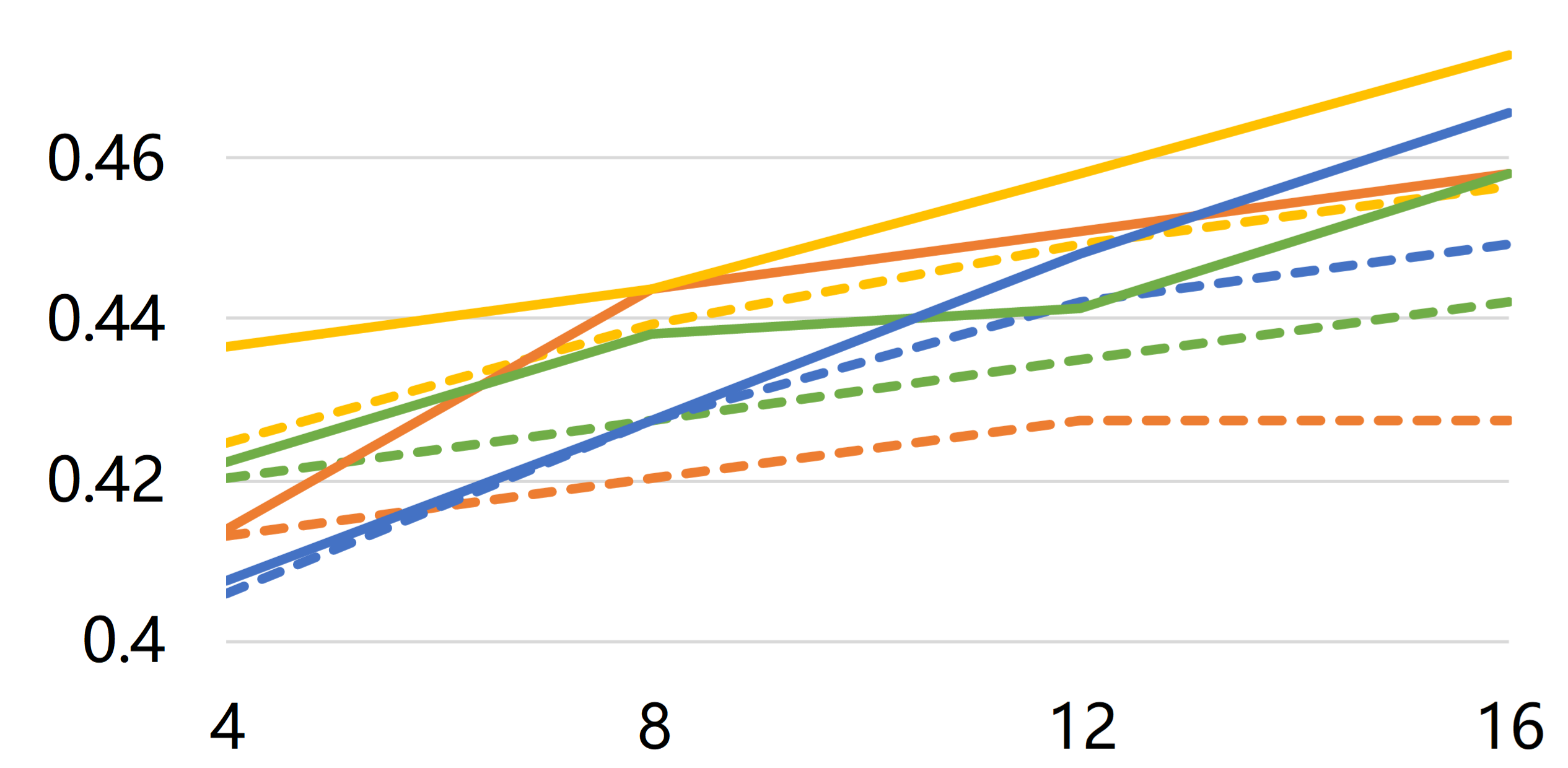}
}
\subfigure[CNC task (Eye)]
{
	\includegraphics[width=0.28\textwidth]{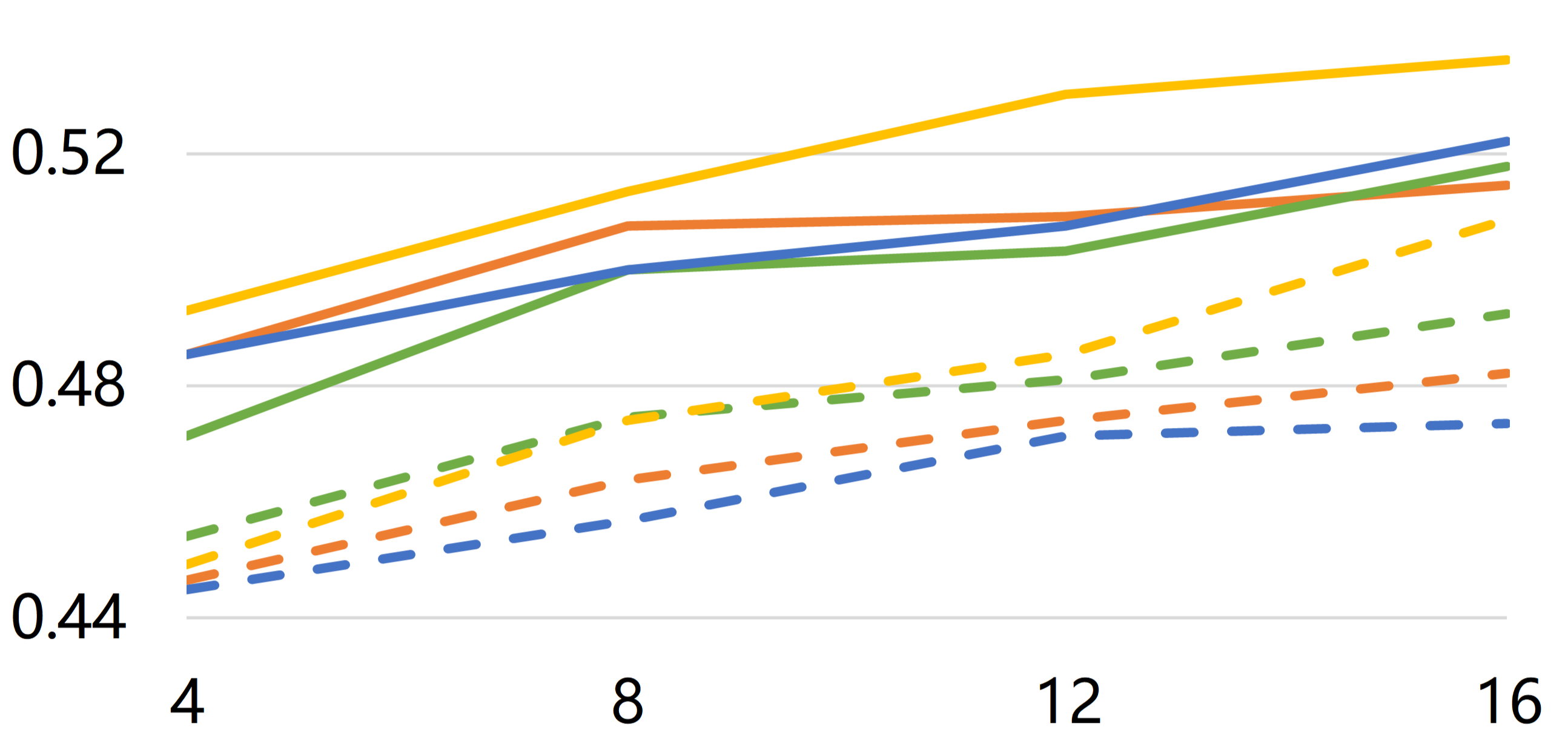}
}
\subfigure[Tense task (Eye)]
{
	\includegraphics[width=0.28\textwidth]{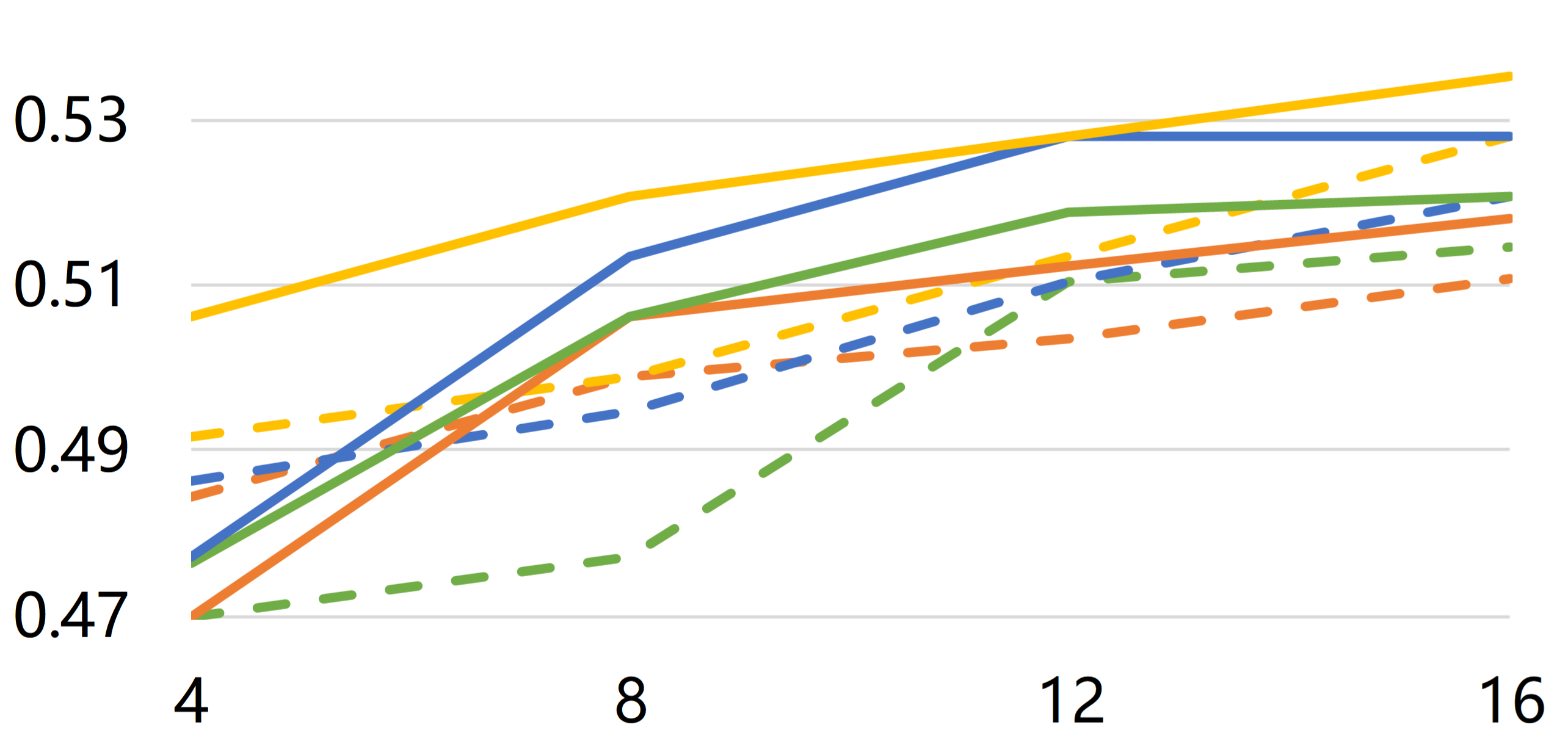}
}
\subfigure[LD task (EEG)]
{
	\includegraphics[width=0.28\textwidth]{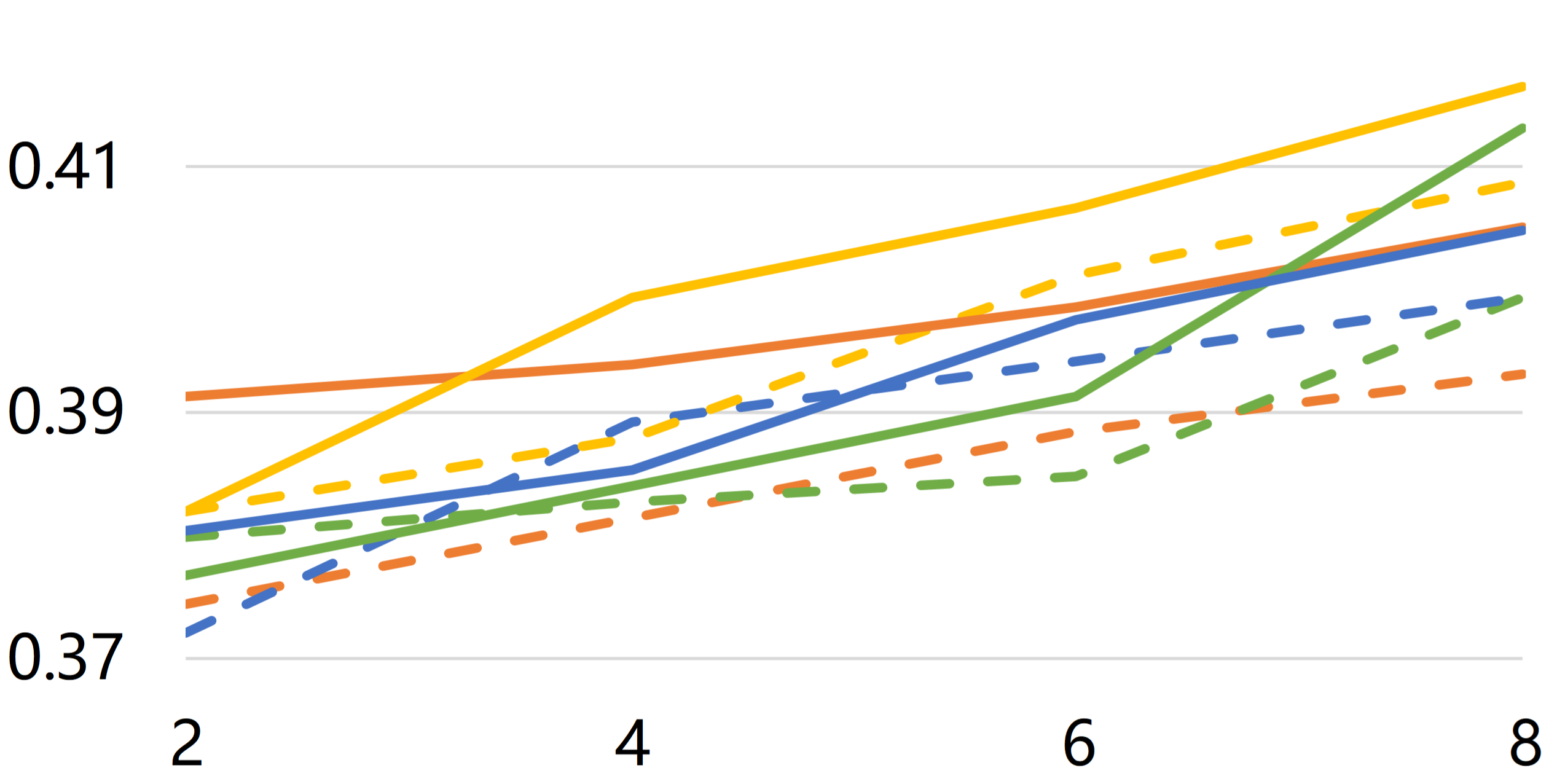}
}
\subfigure[CNC task (EEG)]
{
	\includegraphics[width=0.28\textwidth]{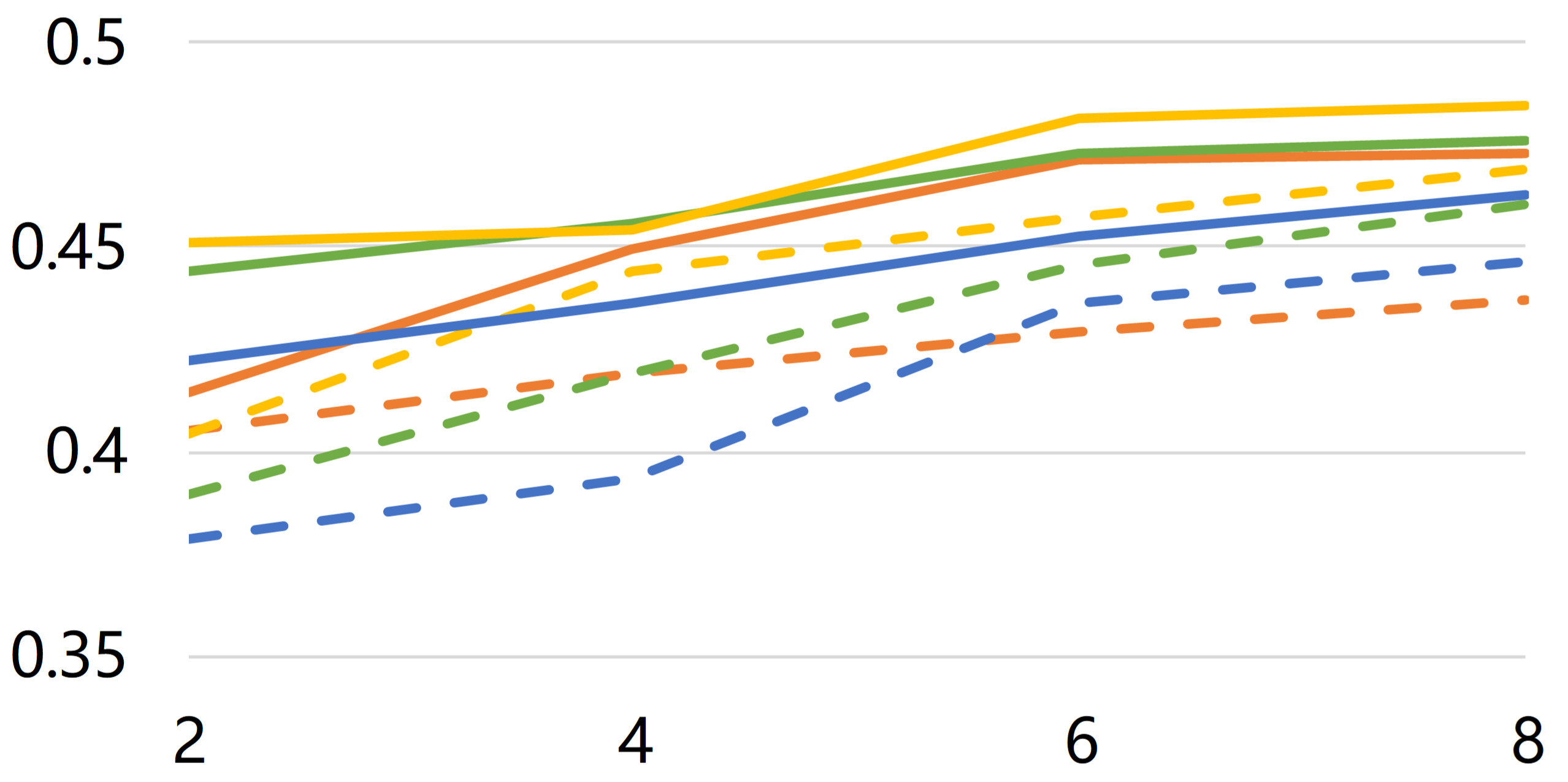}
}
\subfigure[Tense task (EEG)]
{
	\includegraphics[width=0.28\textwidth]{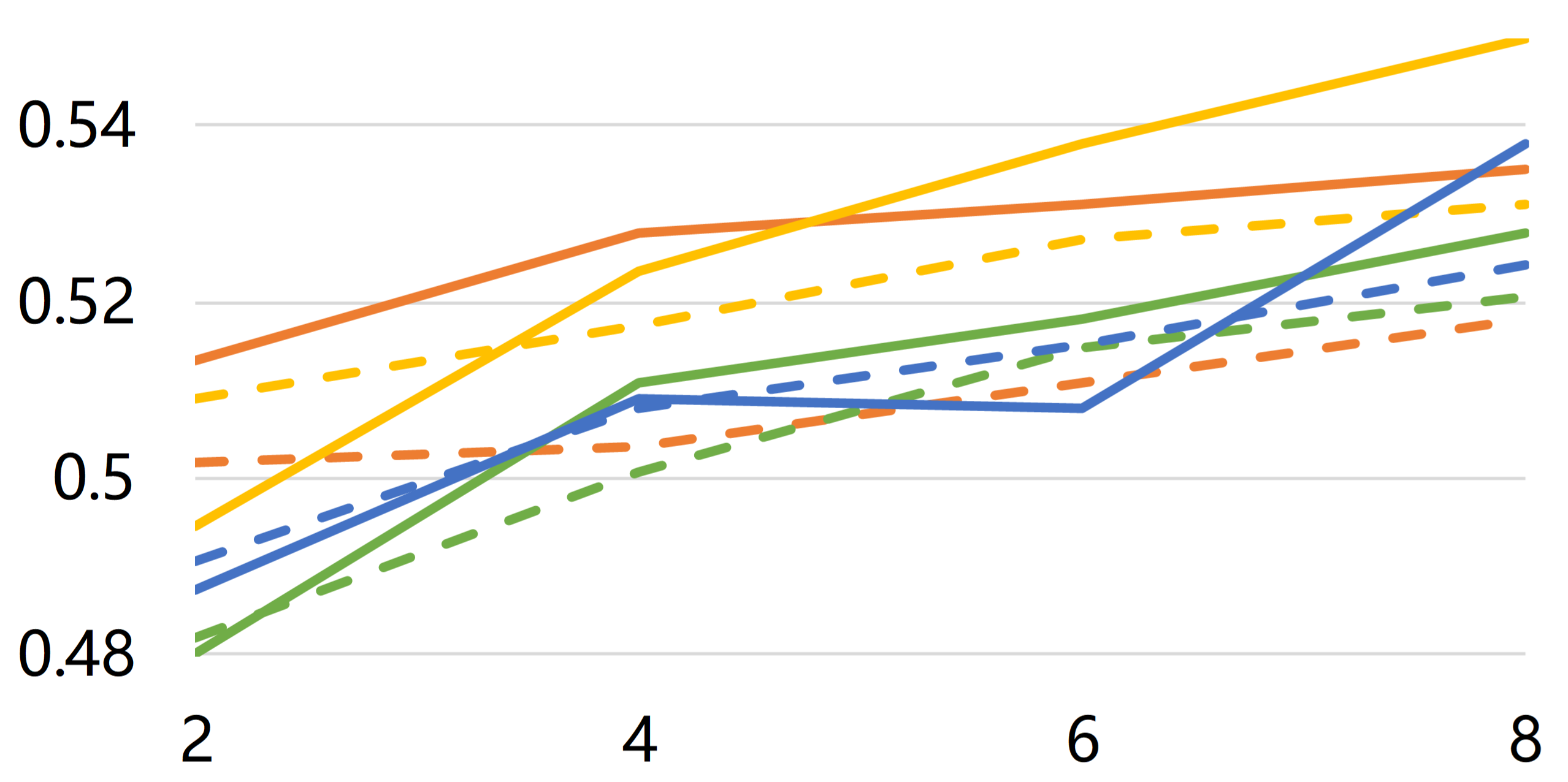}
}
\includegraphics[width=0.85\textwidth]{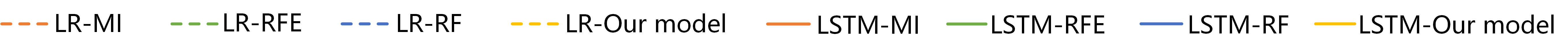}
\caption{Results of our model and other feature selection methods on bridging tasks with eye-tracking features and EEG features. `MI', `RFE' and `RF' denote Mutual Information, Recursive Feature Elimination and Random Forest, respectively. The evaluation metric is $F_1$ (shown in y-coordinate). X-coordinate denotes the number of features selected (i.e., $k$).}\label{fig:3}
\end{figure*}
\subsection{Ablation Study}
To verify the robustness of the attention mechanism in our bridging framework, we further carried out bridging experiments without using the Bi-LSTM encoder. The results for eye-tracking signals are shown in Figure \ref{no Bi-LSTM}, and the results for EEG signals can be found in Appendix. Comparing the complete model to the variant model without the Bi-LSTM encoder, we find that the curves of attention scores obtained by the two models in the same bridging task are in general accord with each other. Although cognitive processing signals with the highest attention scores in some bridging tasks under the two models are different, the overall trends are the same. This suggests that the feature-level attention in our model can well capture alignments between linguistic features and cognitive processing signals.

\subsection{Effectiveness Validation by Signal Masking}
To further validate the effectiveness of our model, we conducted masking experiments on all bridging tasks. First, we sort cognitive processing signals in a descending order according to their attention scores. Then, on the test set, one cognitive processing signal is masked at a time to take a deep look into the role of the attention layer. The results of masking experiments with the two types of cognitive processing signals (eye-tracking and EEG) are shown in Figure \ref{fig:2}. All results are obtained by 5-fold cross validation.

Partially masking some cognitive processing signals results in a drop in performance. Meanwhile, the higher attention score of the cognitive processing signals being masked, the more significant the performance drop, which indicates that the proposed attention layer can measure the degree of importance of features to neural network classification. Additionally, we find that the performance of masking cognitive processing signals with low attention scores is better than the original model that uses all cognitive processing signals. This suggests that some cognitive processing signals may bring negative noises to the bridging task.

\subsection{Comparison to Other Feature Selection Methods}
The attention layer in our framework can be considered as a feature selector: selecting cognitive processing signals that are highly correlated with target linguistic features to feed into the predicting layer. To evaluate classification performance of our attentional feature selection against traditional feature selection methods, we conducted comparative experiments on eye-tracking and EEG signals. We adopted various feature selection methods, including our attention method, to estimate an ``importance'' score for each feature. Then, we selected top $k$ features as inputs to a classifier, and evaluated the trained classifier on the test set to obtain the differences on classification performance.

We used two classification methods to evaluate feature selection: logistic regression classifier (LR) and LSTM model. We chose three different traditional feature selection methods, one from each type of feature selection methods as described in the section of related work: (1) \textbf{Mutual Information}, a filtering method measuring the dependence between two random variables by the joint probability distribution of the two variables and their marginal probability distributions; (2) \textbf{Recursive Feature Elimination}, a wrapper method recursively eliminating  a small number of features that have dependencies and collinearity in the classification model; (3) \textbf{Random Forest}, an embedded method that combines a number of decision tree classifiers and returns average prediction results of these classifiers. We conducted experiments on all bridging tasks with 5-fold cross-validation. The feature importance scores estimated by these methods and our model on bridging tasks (LD, CNC and Tense) are shown in Table \ref{Tab:ranking}.

Due to the space limit, we only show partial results in Figure \ref{fig:3} while the full results can be found in Appendix. Obviously, either with LR or LSTM, the classification performance with features selected by our method is significantly better than that with features selected by the three traditional methods on almost all bridging tasks. It suggests that our proposed method is able to effectively select useful features. In addition, the performance of the LSTM model is better than LR model in most cases. However, LR outperforms LSTM on some tasks when the number of selected features is small. We conjecture that this may be due to the overfitting problem of LSTM on small training sets with less feature information.

\section{Conclusions}
In this paper, we have presented a unified model to investigate the relationship between cognitive processing signals and linguistic features by using a feature-level attention mechanism to select relevant cognitive processing signals in linguistic bridging tasks. This method can test a wide variety of linguistic features on a single dataset with cognitive processing signals recorded under natural reading. Such a data-driven method may act as a surrogate to controlled experiments in neurolinguistics. Experiment results corroborate and extend previous findings, demonstrating that our method can effectively detect linguistic information in cognitive processing signals.

\section*{Acknowledgments}
The present research was supported by Zhejiang Lab (No. 2022KH0AB01) and the Natural Science Foundation of Tianjin (No. 19JCZDJC31400). We would like to thank the anonymous reviewers for their insightful comments.

\bibliography{aaai22}
\clearpage
\appendix
\section{Appendix}
\begin{figure*}[ht]
\centering
\subfigure[Lexical-EYE bridging tasks)]
{
	\includegraphics[width=0.3\textwidth]{eye-none-lexical.png}
}
\subfigure[Syntactic-EYE bridging tasks]
{
	\includegraphics[width=0.3\textwidth]{eye-none-syntactic.png}
}
\subfigure[Semantic-EYE bridging tasks]
{
	\includegraphics[width=0.3\textwidth]{eye-none-semantic.png}
}
\subfigure[Lexical-EEG bridging tasks]
{
	\includegraphics[width=0.3\textwidth]{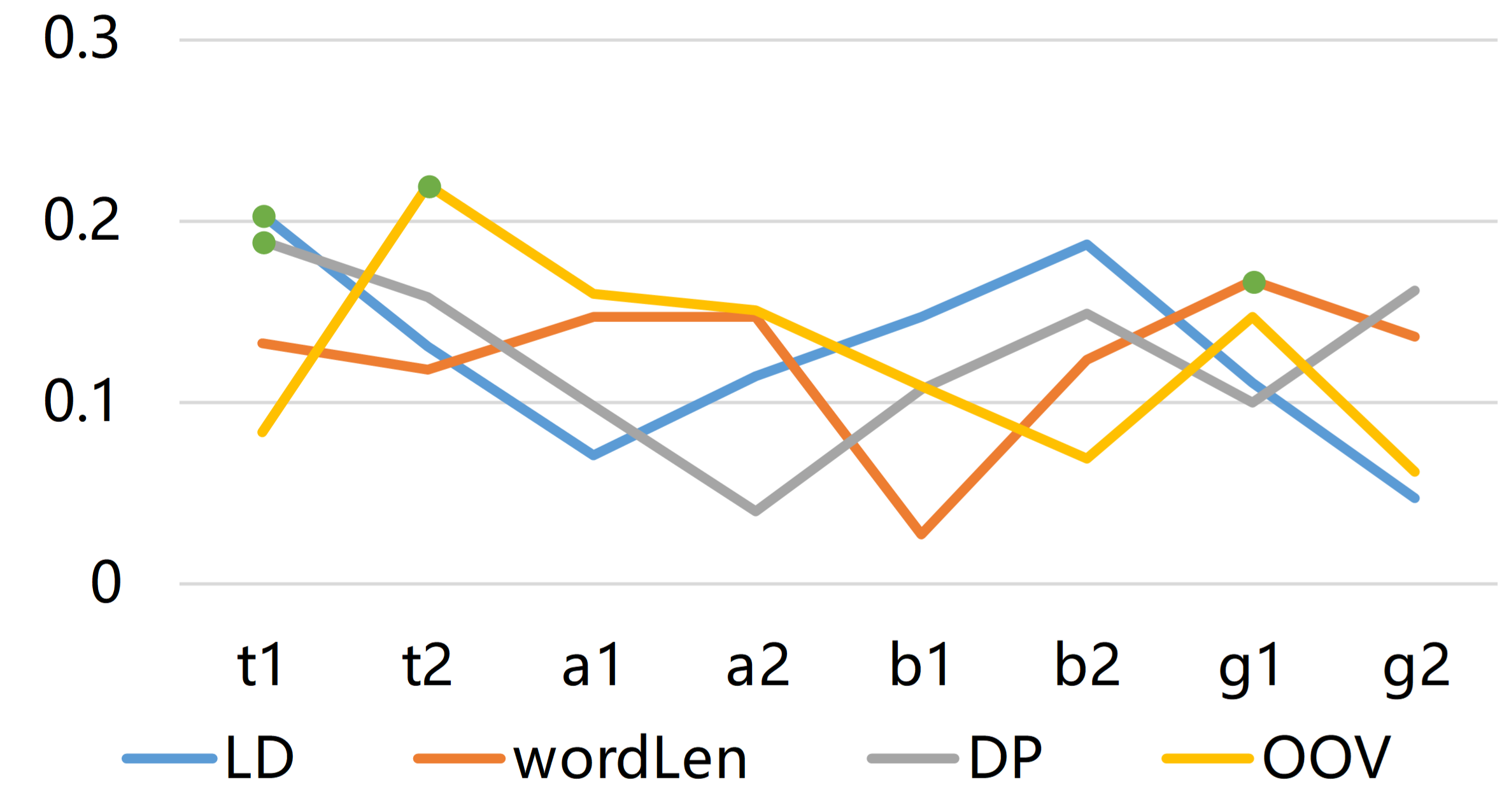}\label{fig:eeg-none-lexical}
}
\subfigure[Syntactic-EEG bridging tasks]
{
	\includegraphics[width=0.3\textwidth]{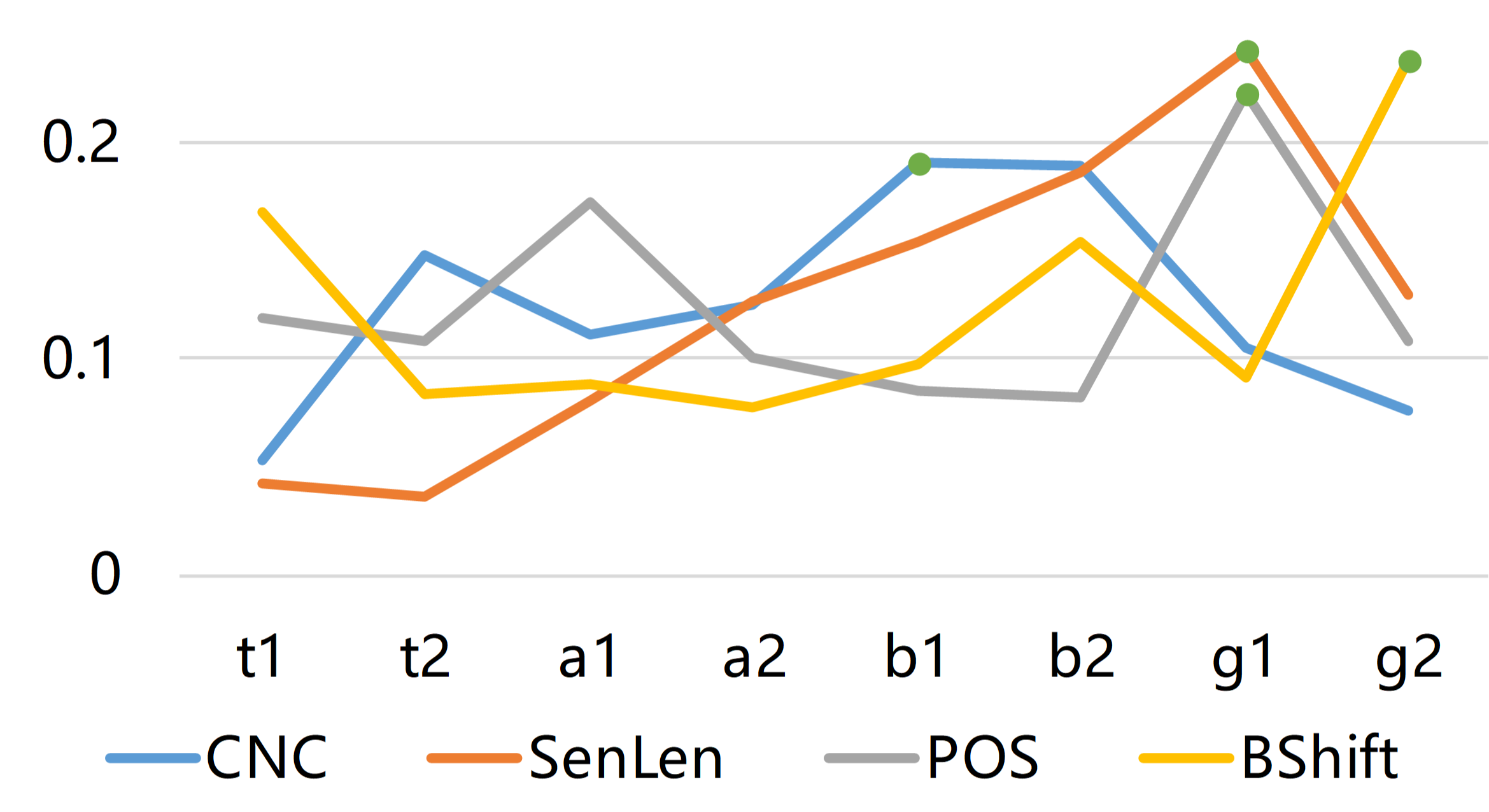}
}
\subfigure[Semantic-EEG bridging tasks]
{
	\includegraphics[width=0.3\textwidth]{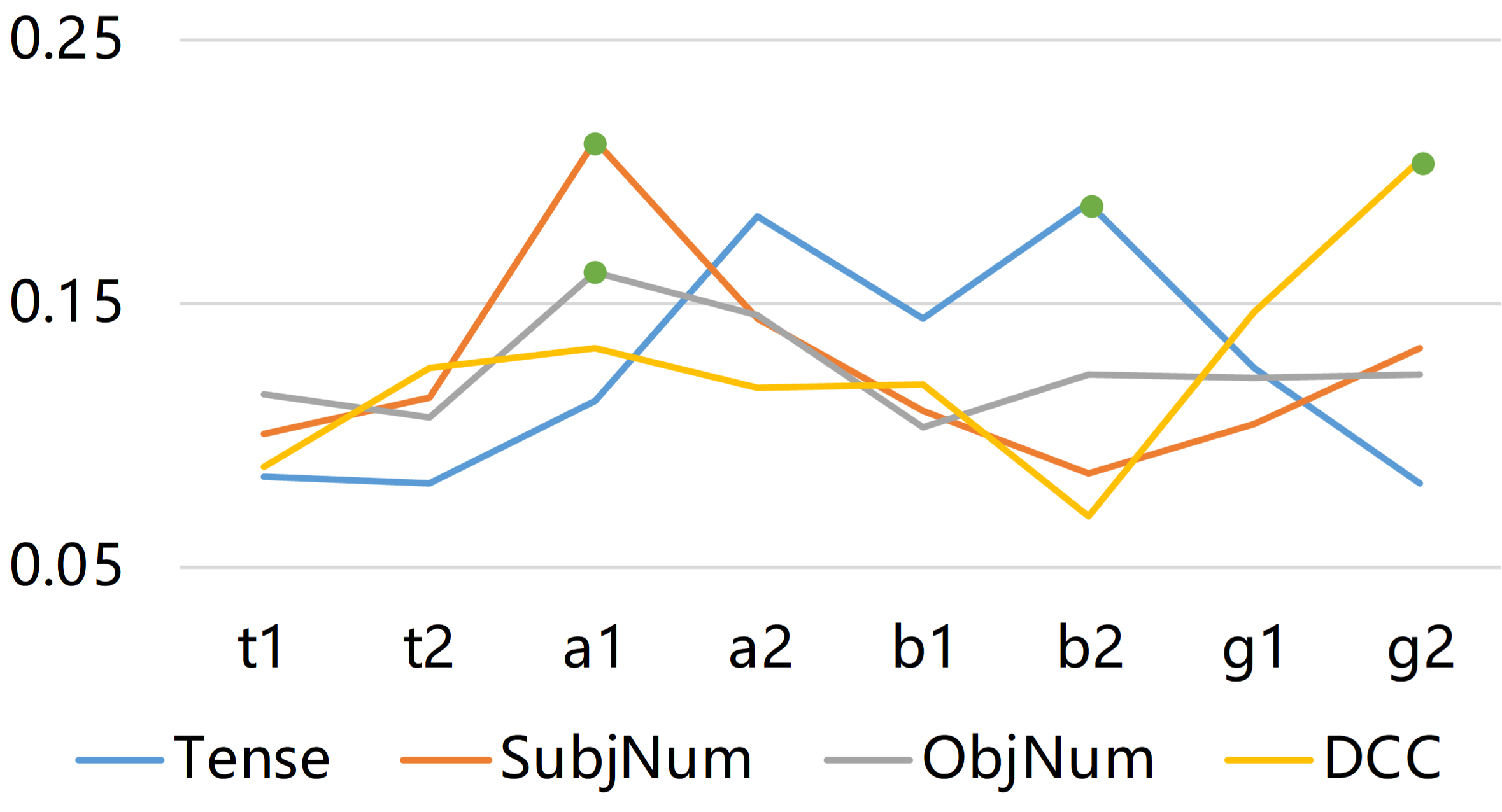}
}
\caption{Complete results of bridging between cognitive processing signals and linguistic features with the variant model that does not use the Bi-LSTM encoder.}\label{appendix:1}
\end{figure*}
\begin{figure*}[ht]
\centering
\subfigure[LD task]
{
	\includegraphics[width=0.23\textwidth]{eye-ld-com.png}
}
\subfigure[WordLen task]
{
	\includegraphics[width=0.23\textwidth]{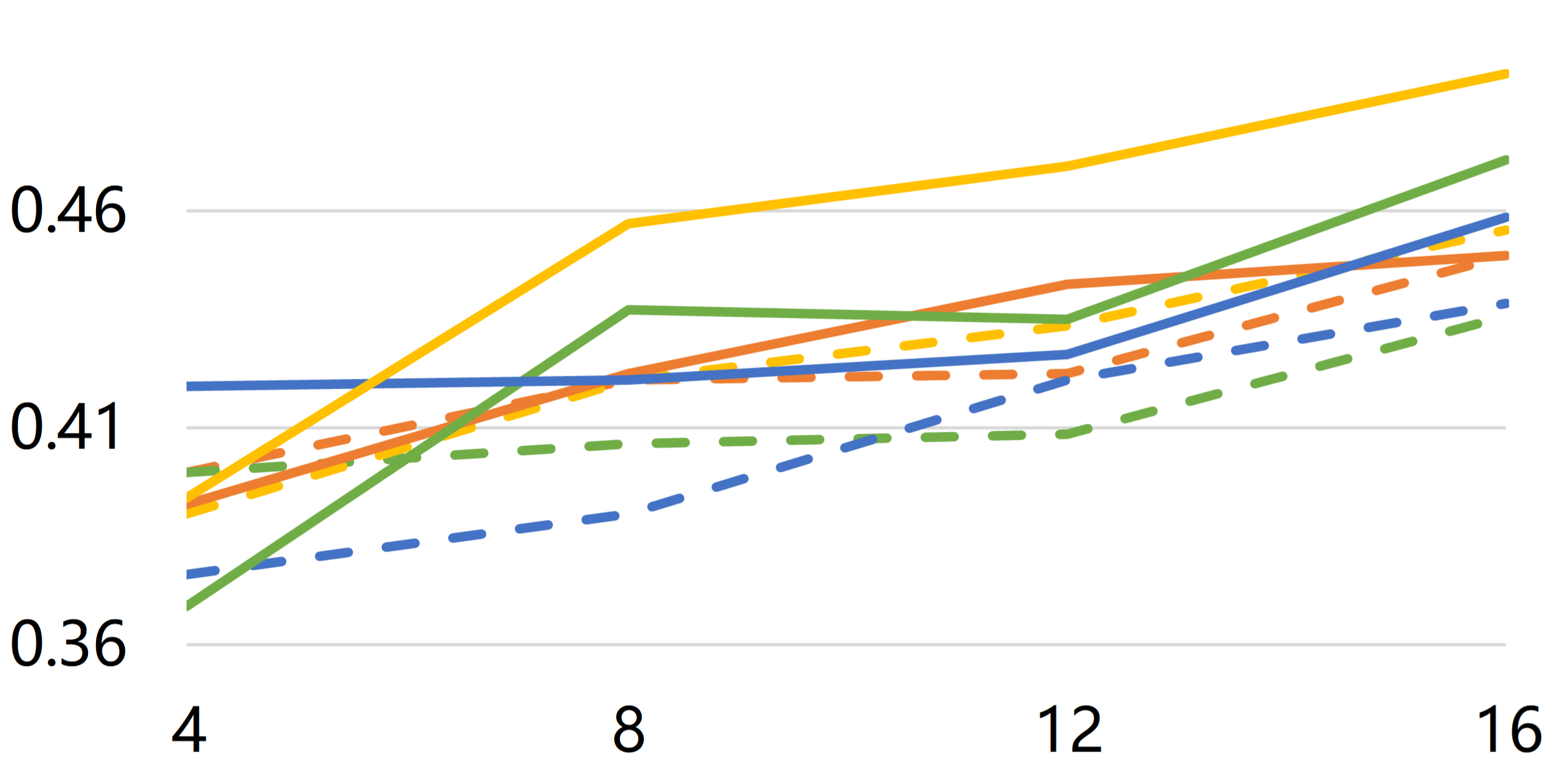}
}
\subfigure[DP task]
{
	\includegraphics[width=0.23\textwidth]{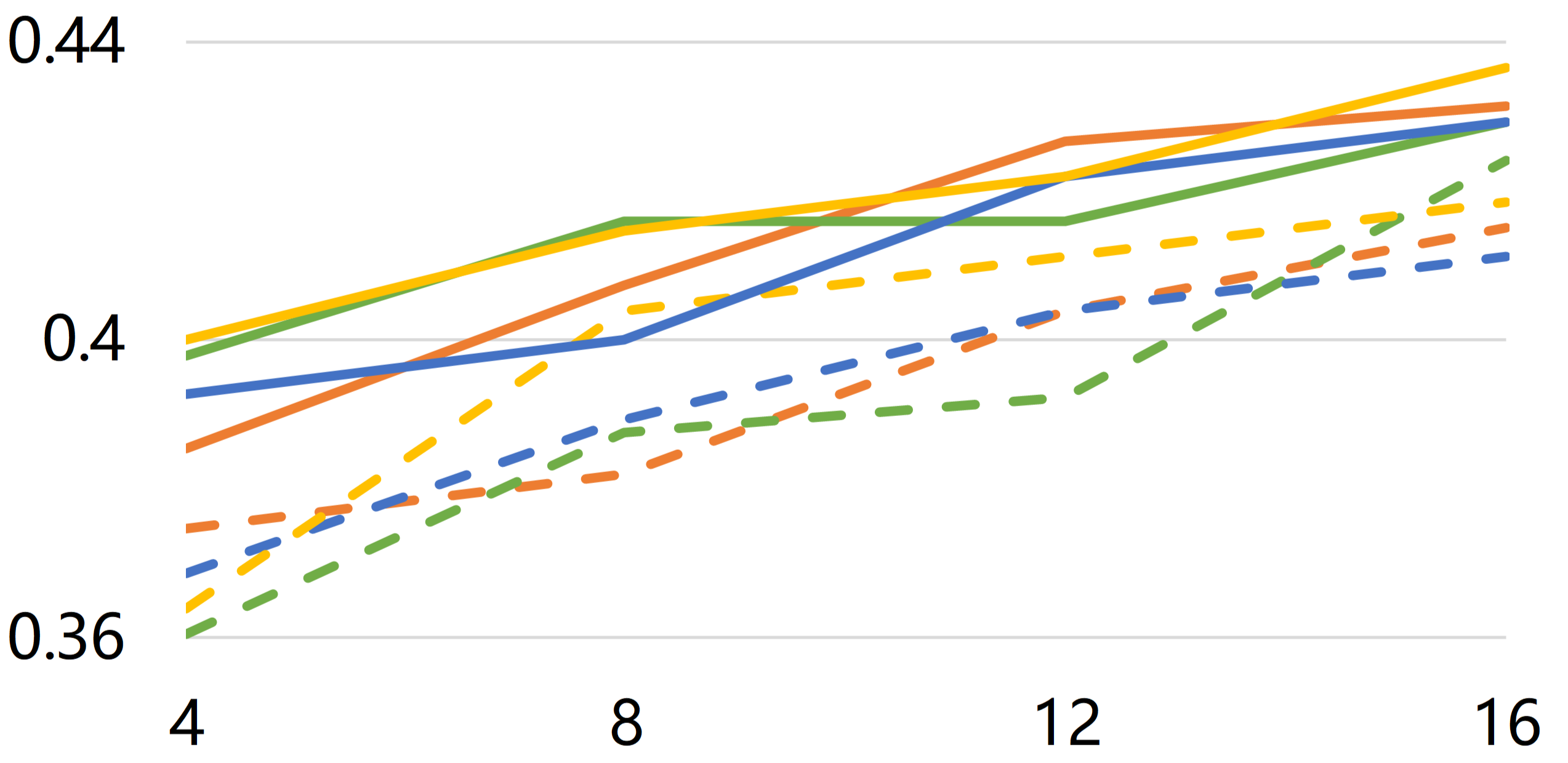}
}
\subfigure[OOV task]
{
	\includegraphics[width=0.23\textwidth]{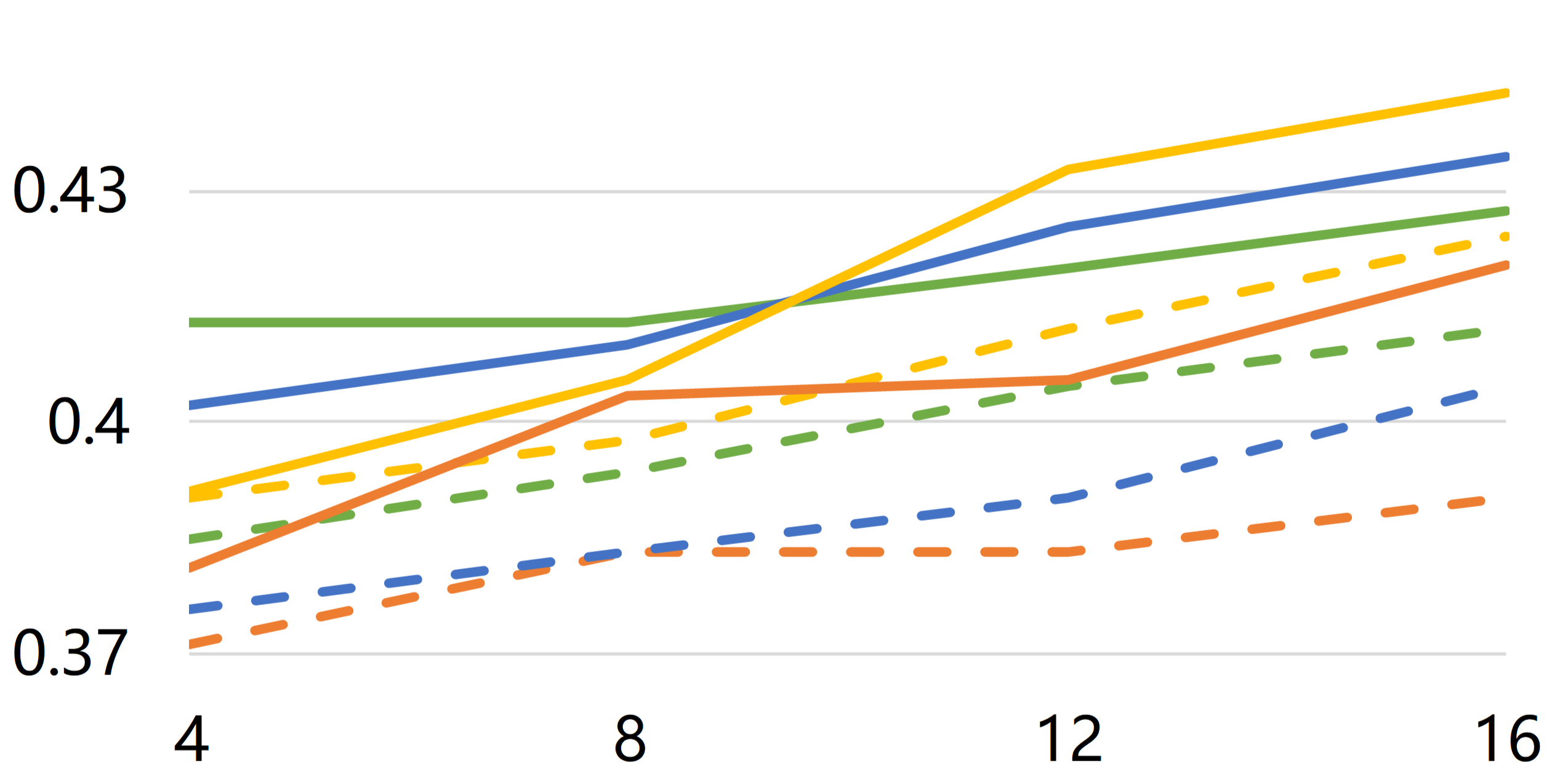}
}
\subfigure[CNC task]
{
	\includegraphics[width=0.23\textwidth]{eye-cnc-com.png}
}
\subfigure[SenLen task]
{
	\includegraphics[width=0.23\textwidth]{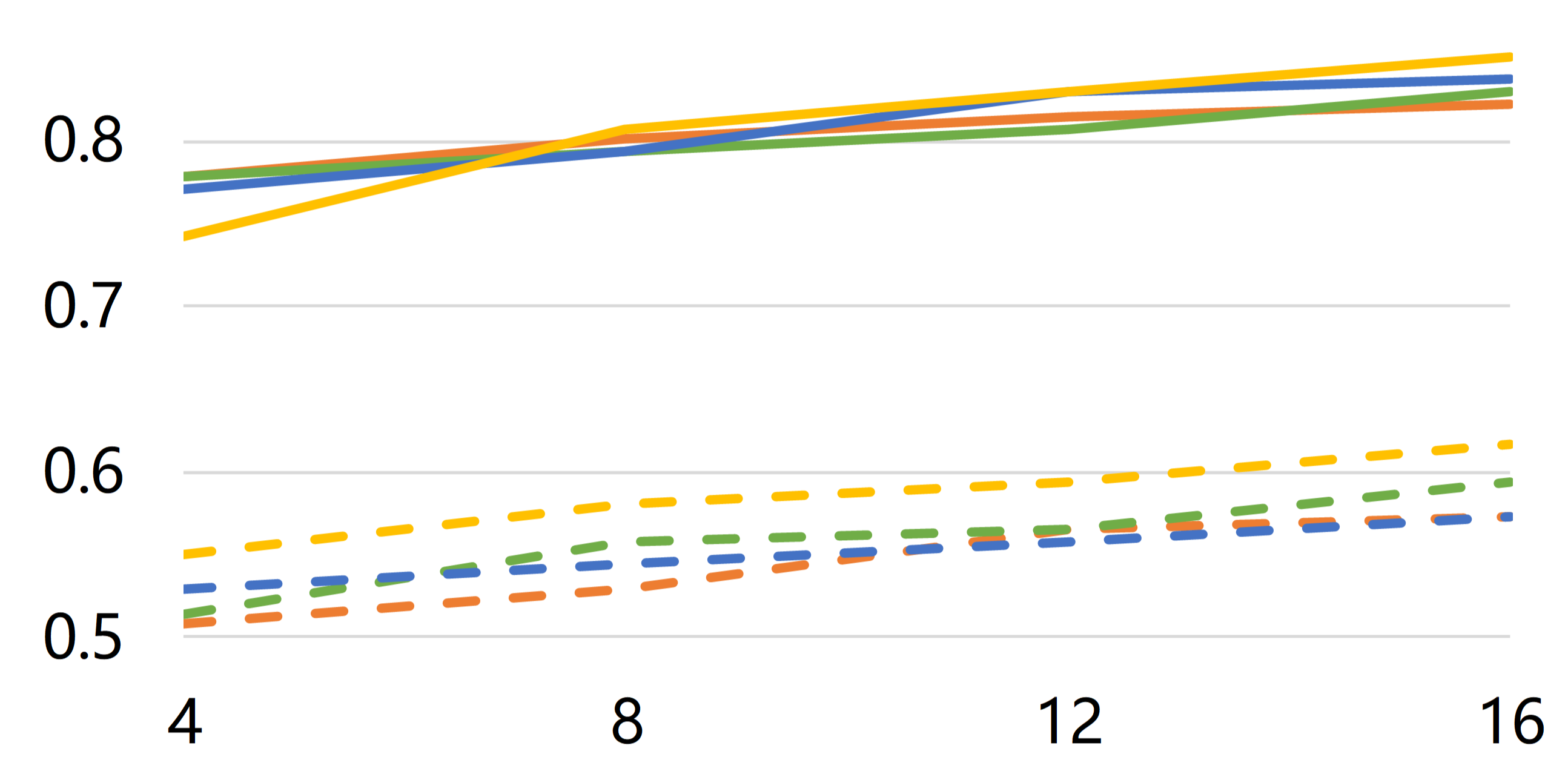}
}
\subfigure[BShift task]
{
	\includegraphics[width=0.23\textwidth]{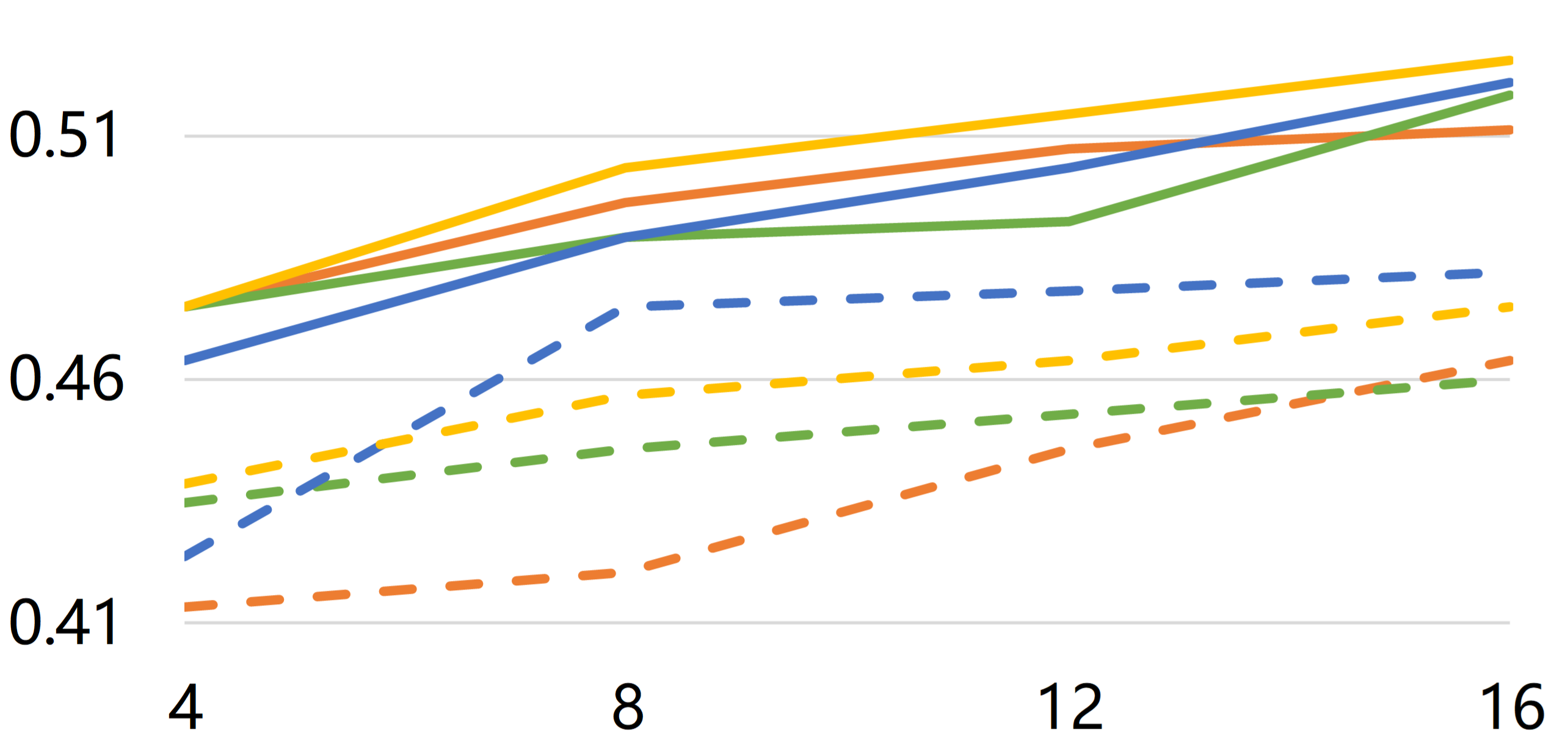}
}
\subfigure[Tense task]
{
	\includegraphics[width=0.23\textwidth]{eye-tense-com.png}
}
\subfigure[SubjNum task]
{
	\includegraphics[width=0.23\textwidth]{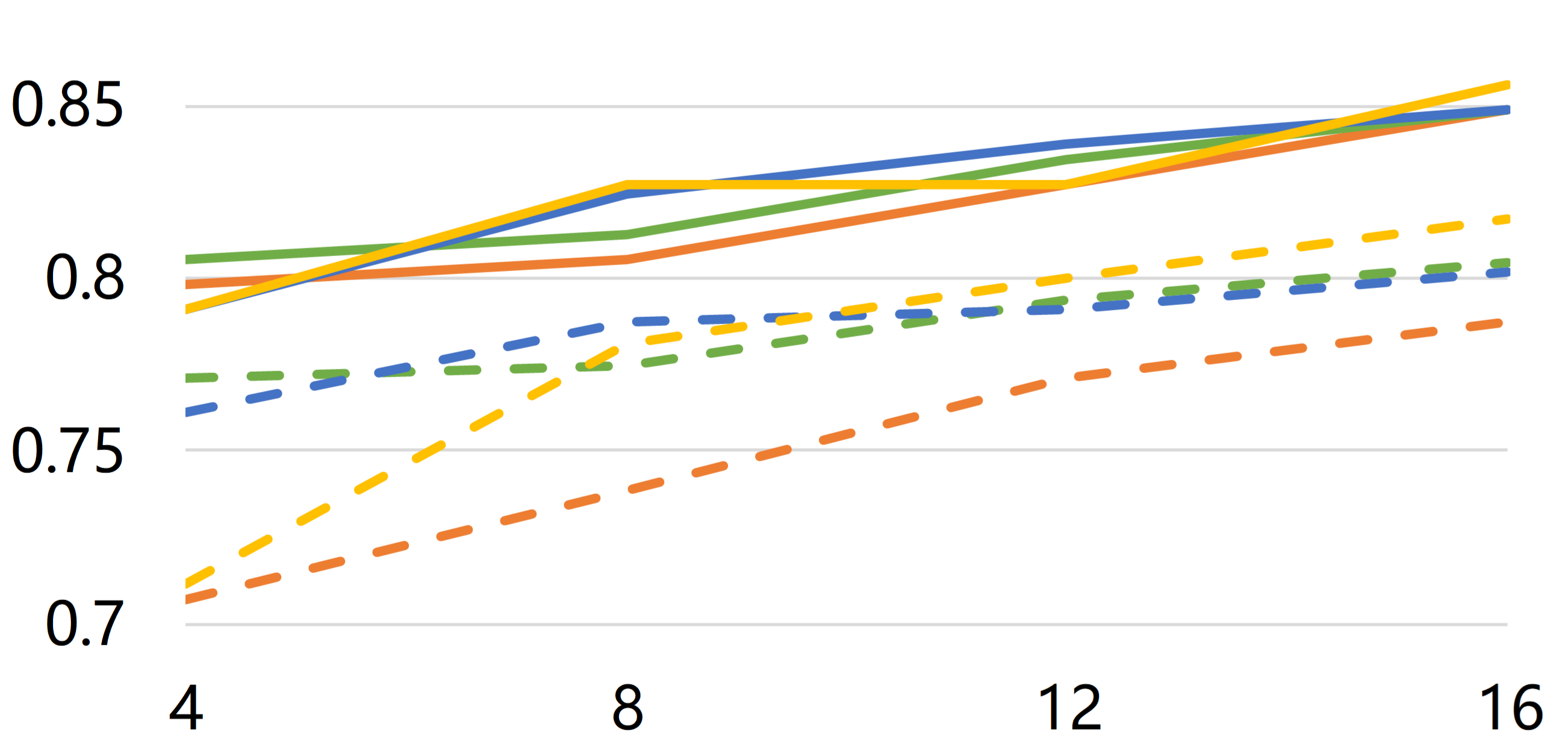}
}
\subfigure[ObjNum task]
{
	\includegraphics[width=0.23\textwidth]{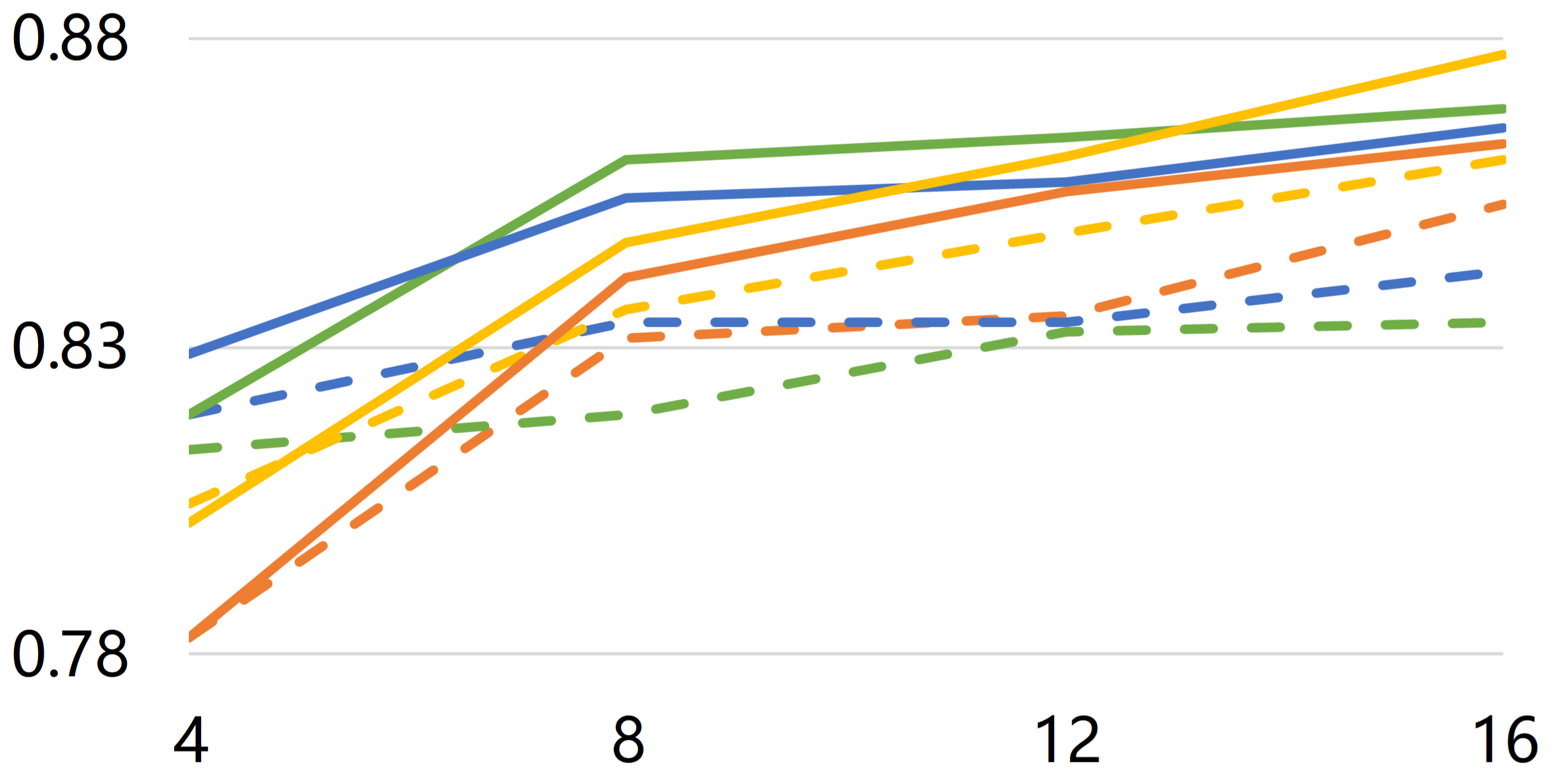}
}
\subfigure[DCC task]
{
	\includegraphics[width=0.23\textwidth]{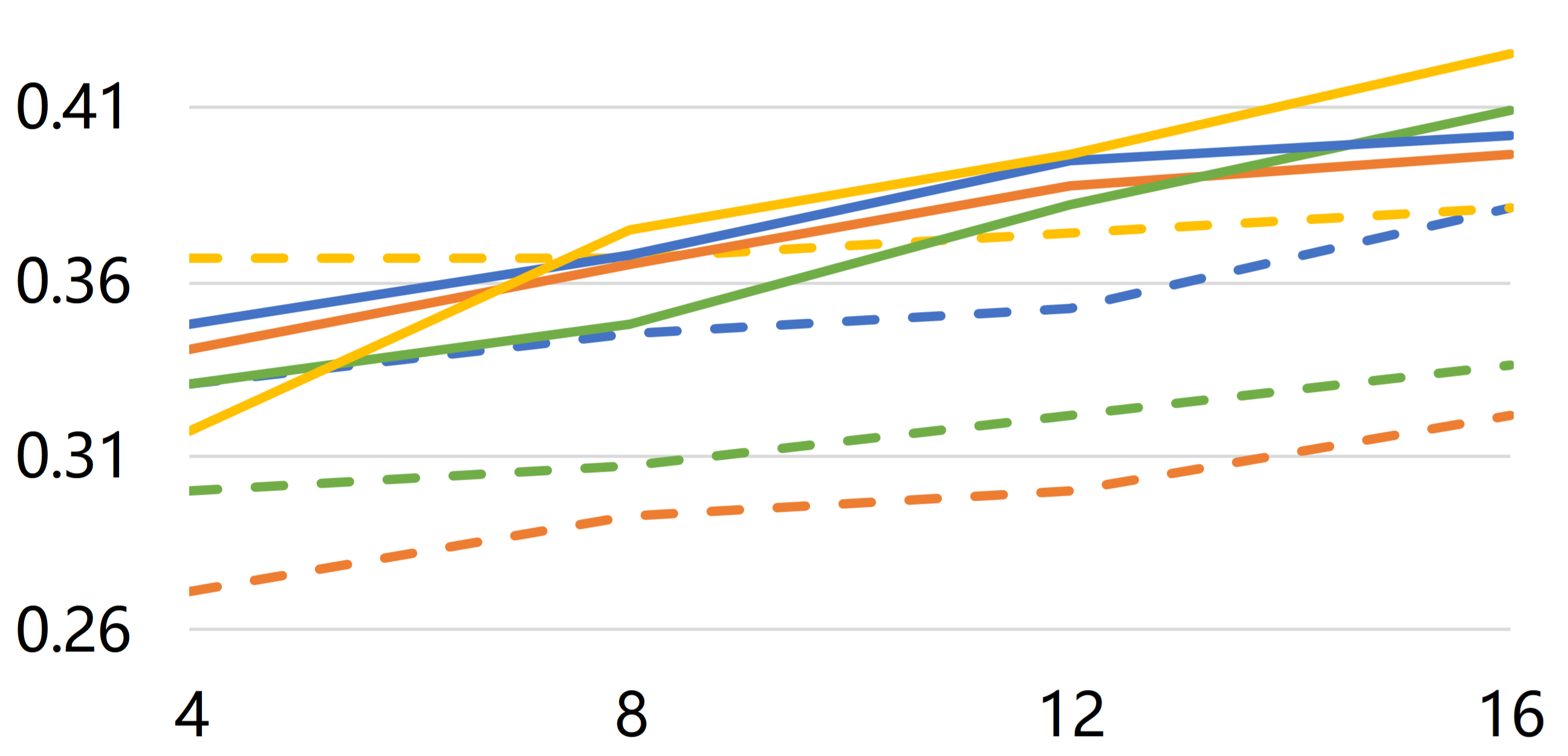}
}
\includegraphics[width=0.85\textwidth]{tuli.png}
\caption{Results of our model and other feature selection methods on bridging tasks with eye-tracking features.}\label{appendix:2}
\end{figure*}
\begin{figure*}[ht]
\centering
\subfigure[LD task]
{
	\includegraphics[width=0.3\textwidth]{eeg-ld-com.png}
}
\subfigure[WordLen task]
{
	\includegraphics[width=0.3\textwidth]{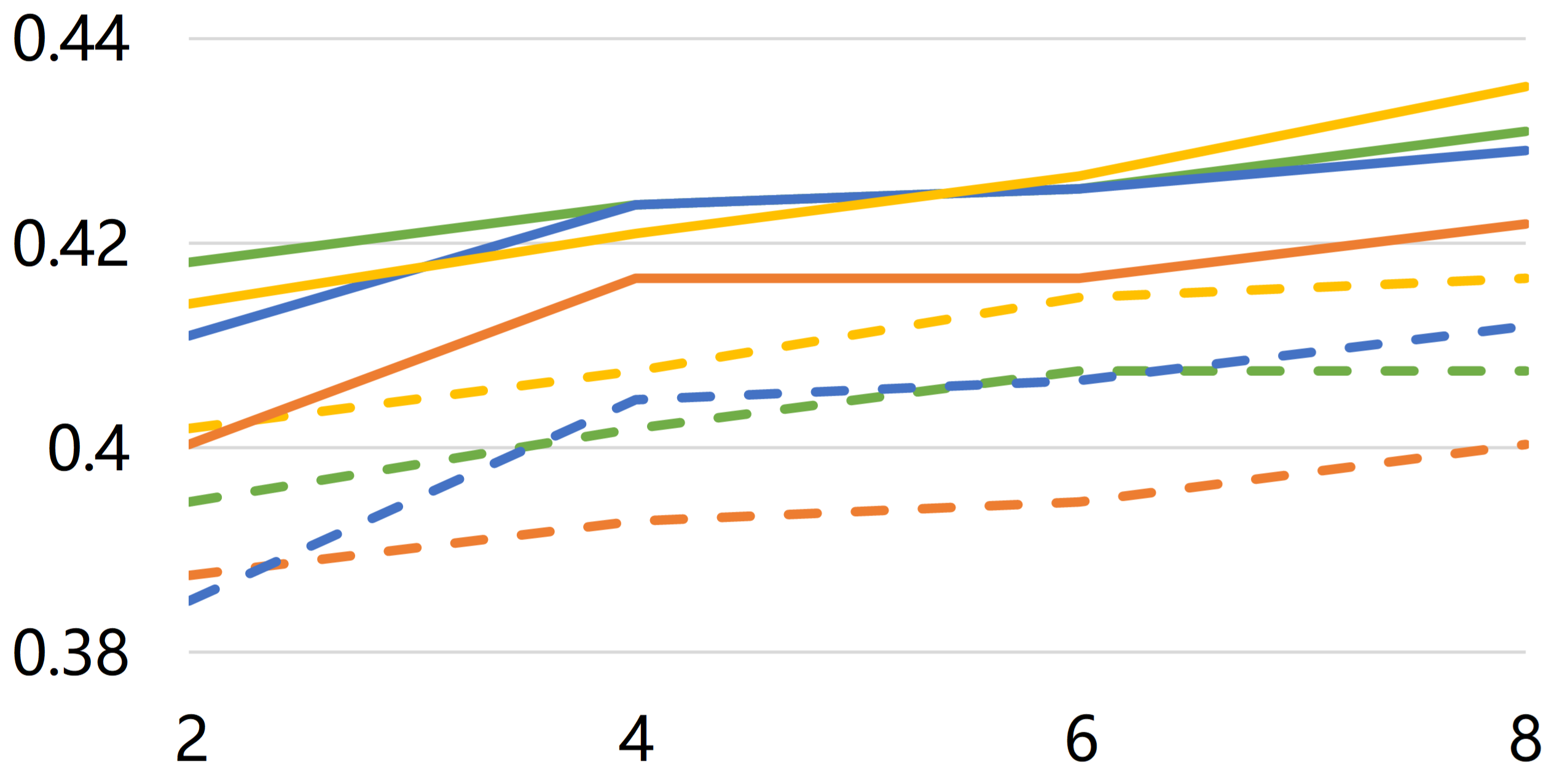}
}
\subfigure[DP task]
{
	\includegraphics[width=0.3\textwidth]{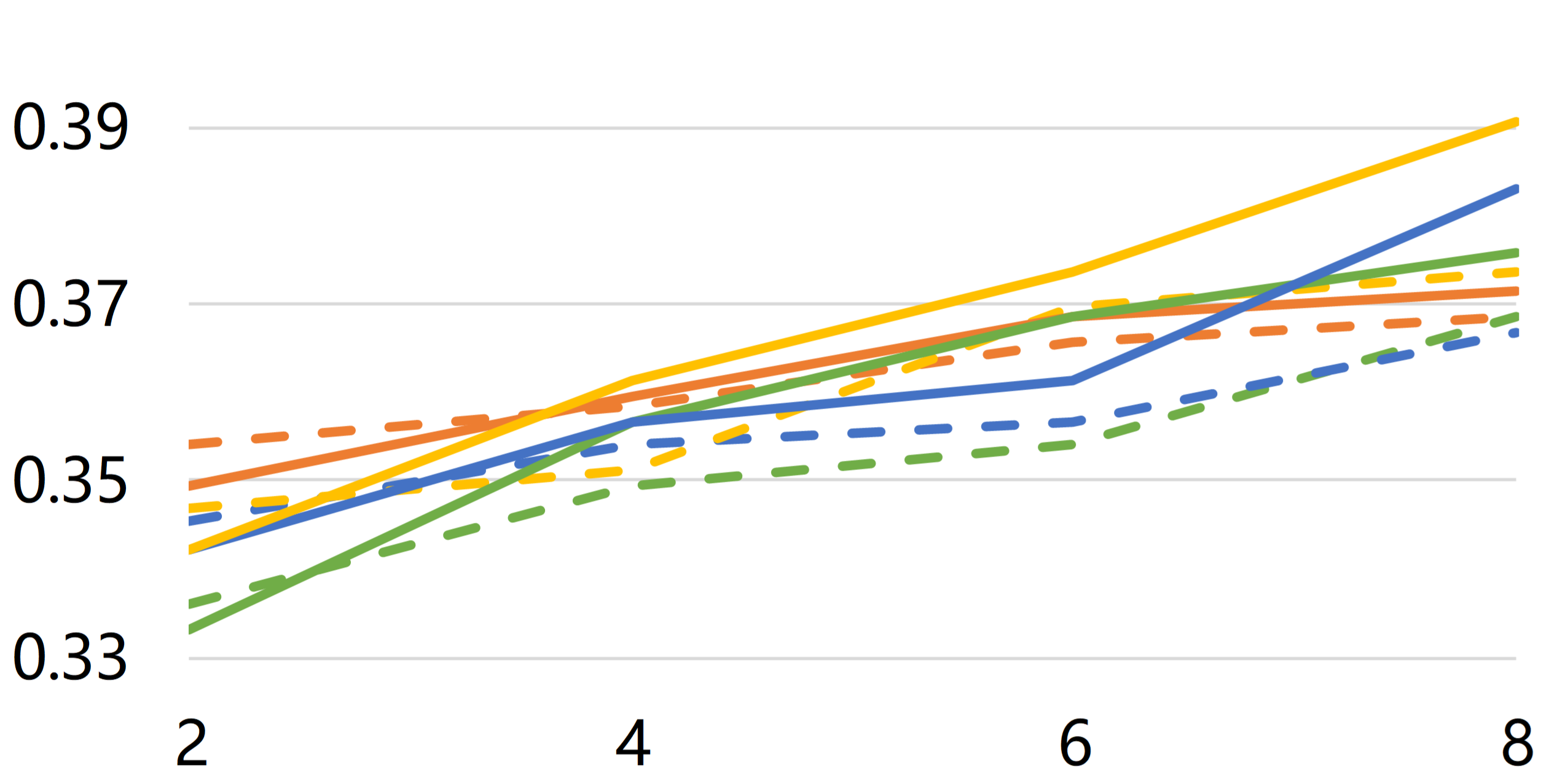}
}
\subfigure[OOV task]
{
	\includegraphics[width=0.3\textwidth]{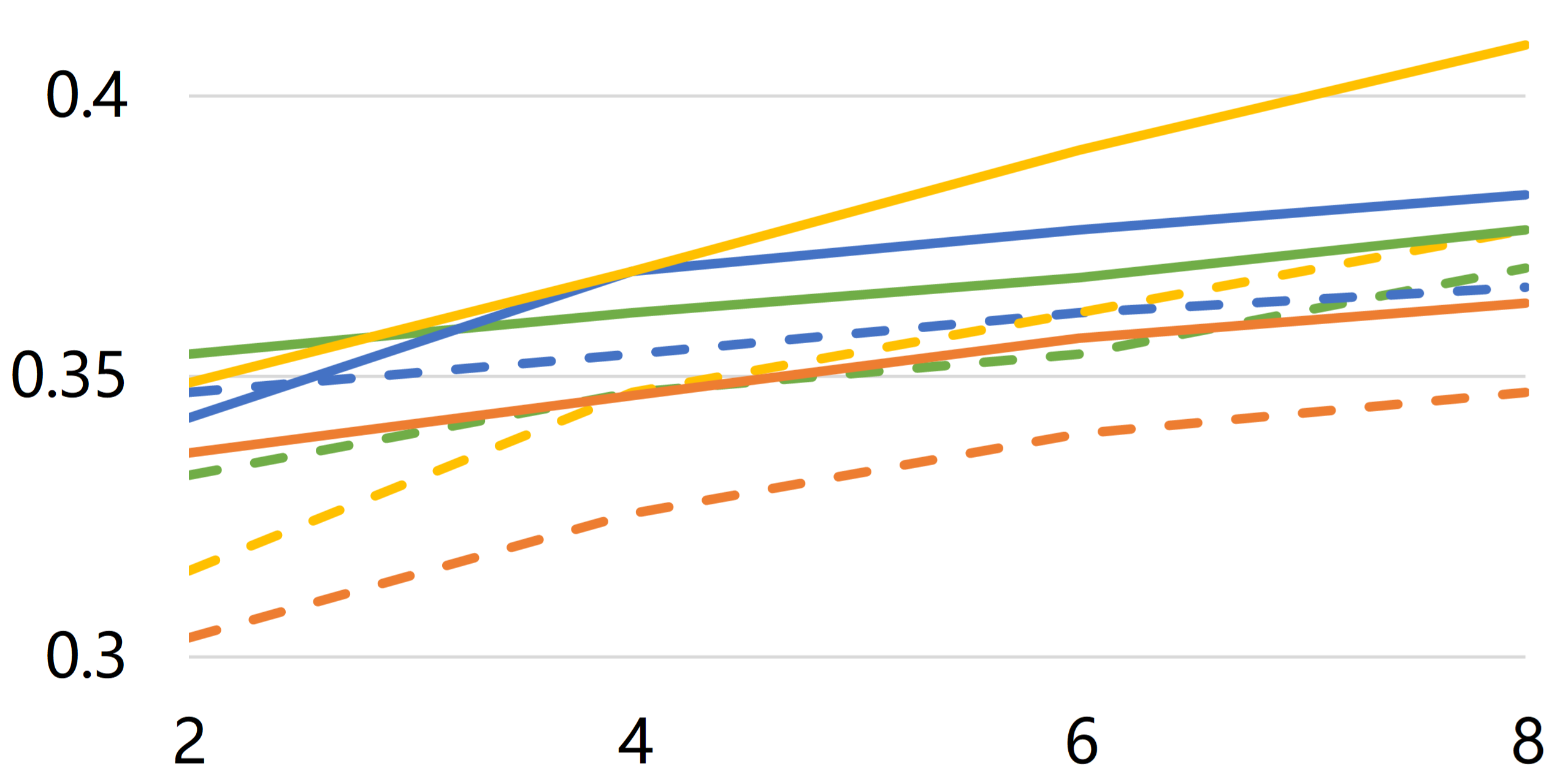}
}
\subfigure[CNC task]
{
	\includegraphics[width=0.3\textwidth]{eeg-cnc-com.png}
}
\subfigure[SenLen task]
{
	\includegraphics[width=0.3\textwidth]{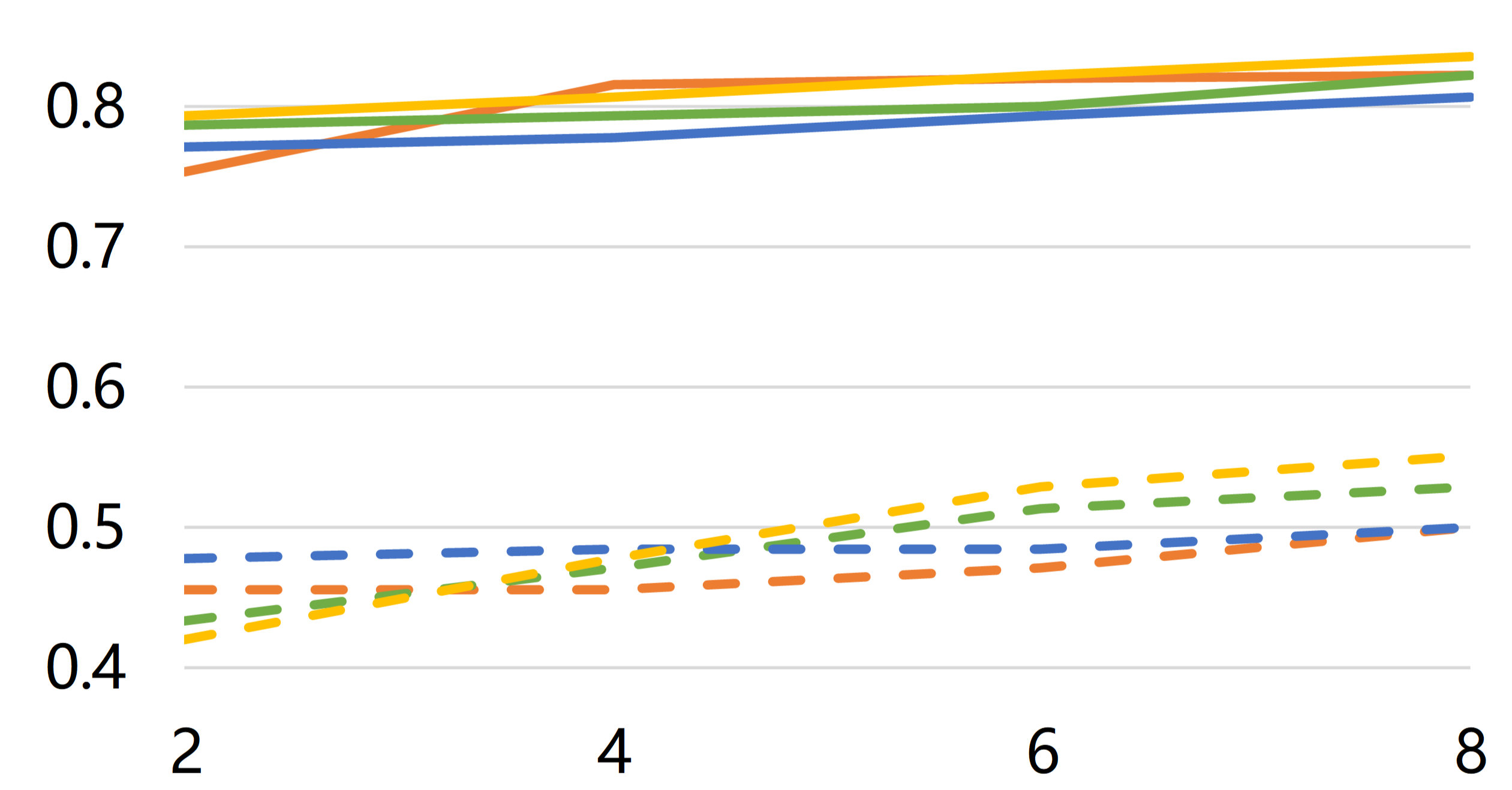}
}
\subfigure[BShift task]
{
	\includegraphics[width=0.3\textwidth]{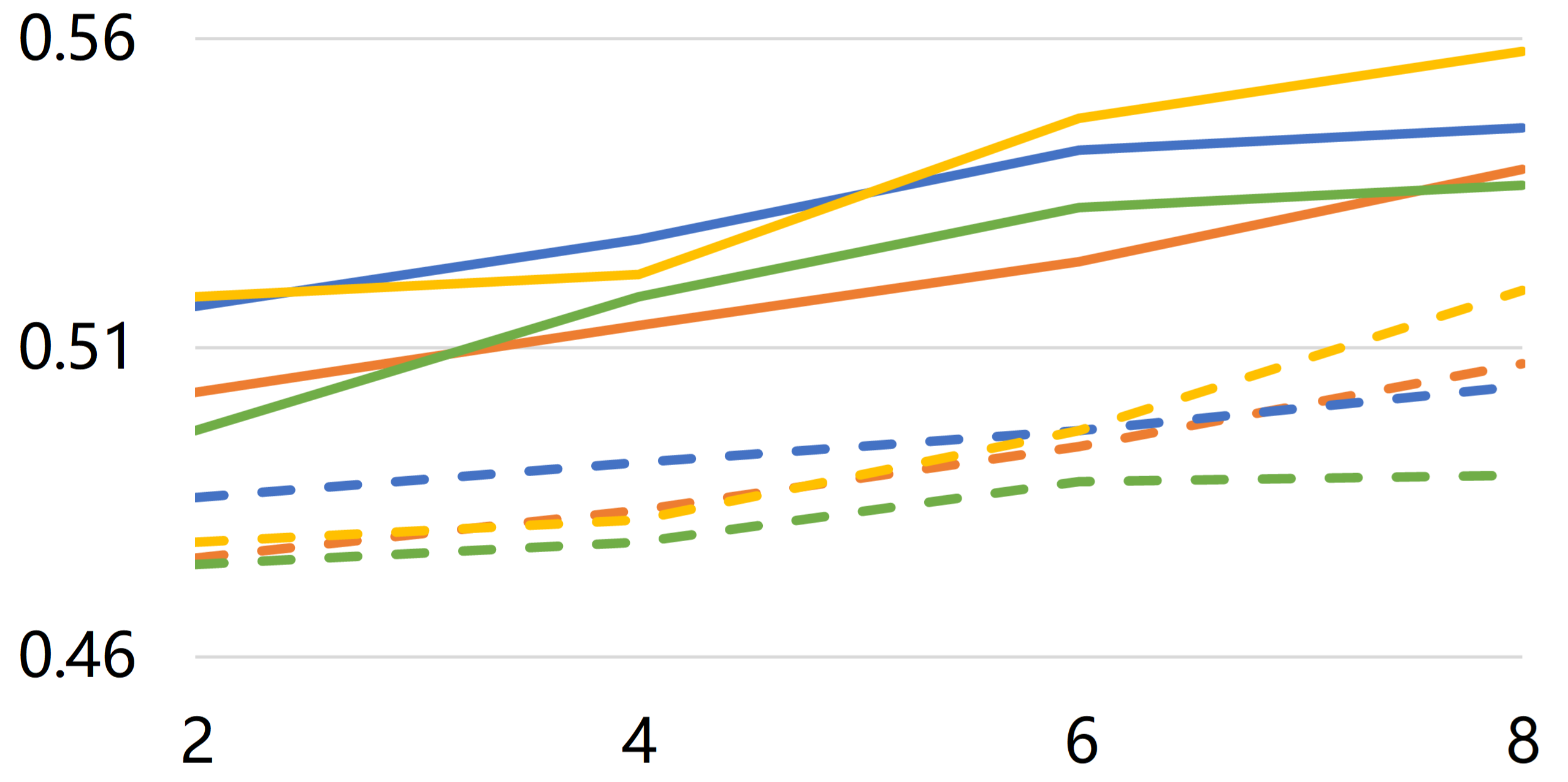}
}
\subfigure[Tense task]
{
	\includegraphics[width=0.3\textwidth]{eeg-tense-com.png}
}
\subfigure[SubjNum task]
{
	\includegraphics[width=0.3\textwidth]{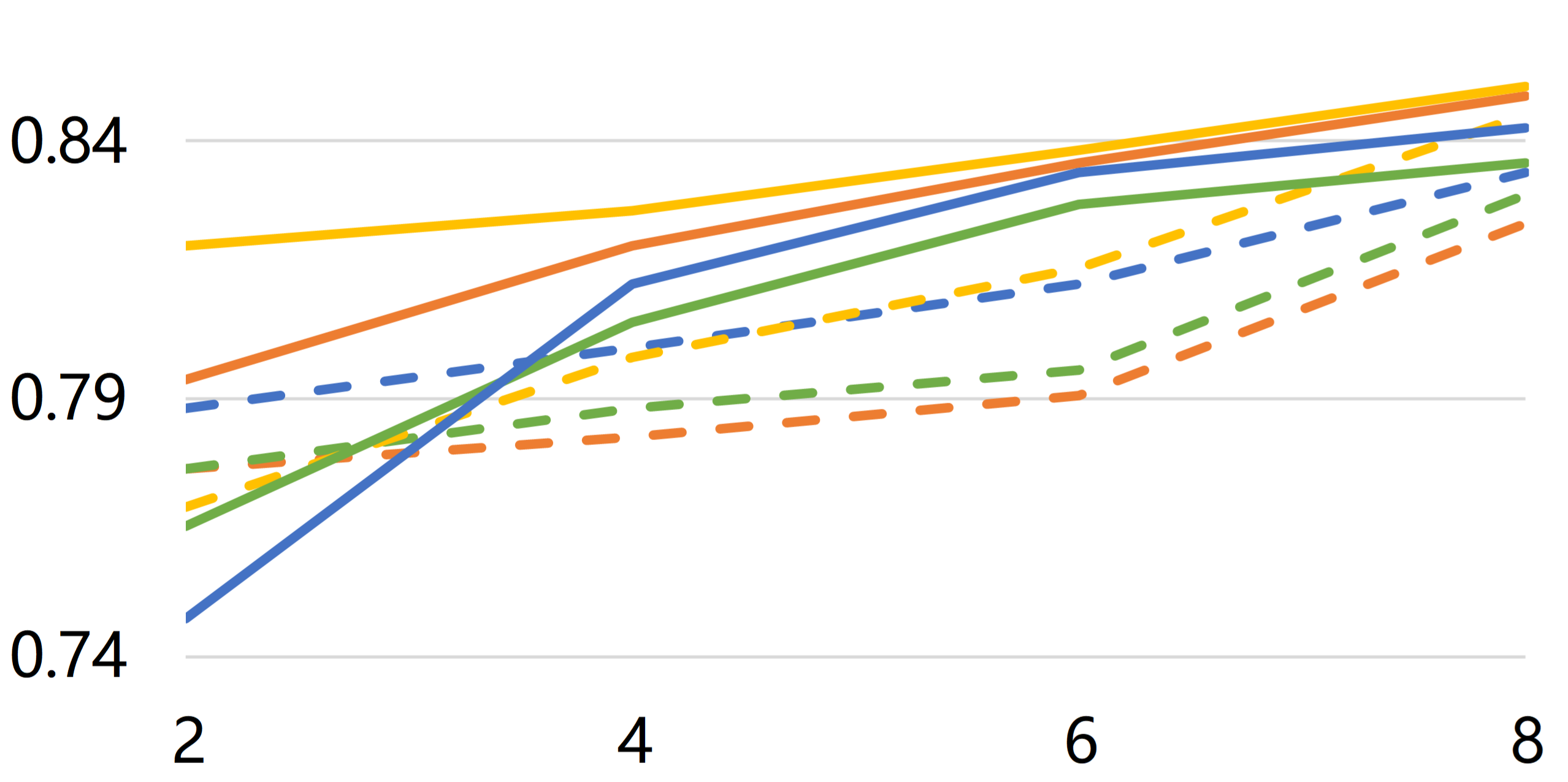}
}
\subfigure[ObjNum task]
{
	\includegraphics[width=0.3\textwidth]{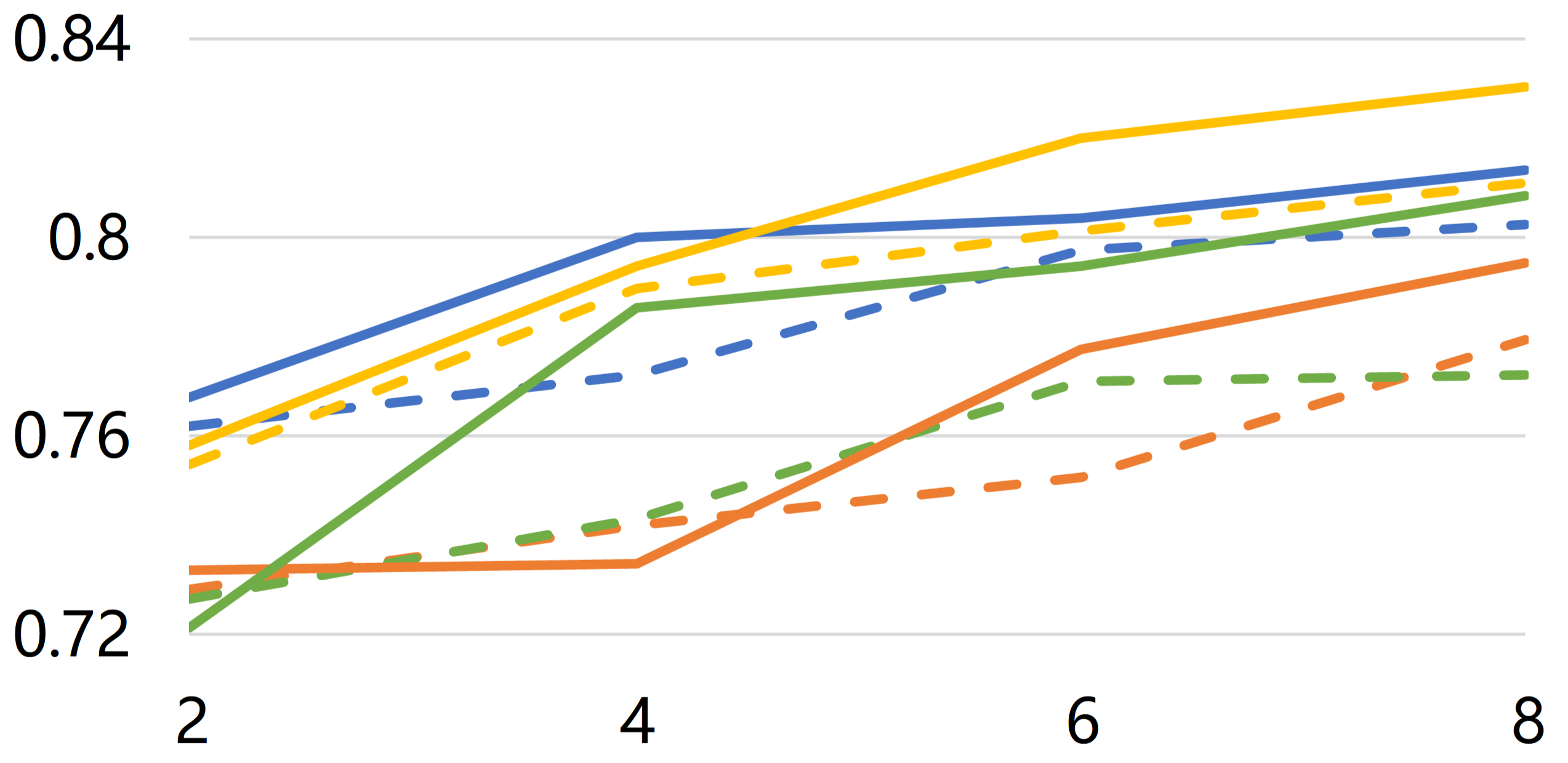}
}
\subfigure[DCC task]
{
	\includegraphics[width=0.3\textwidth]{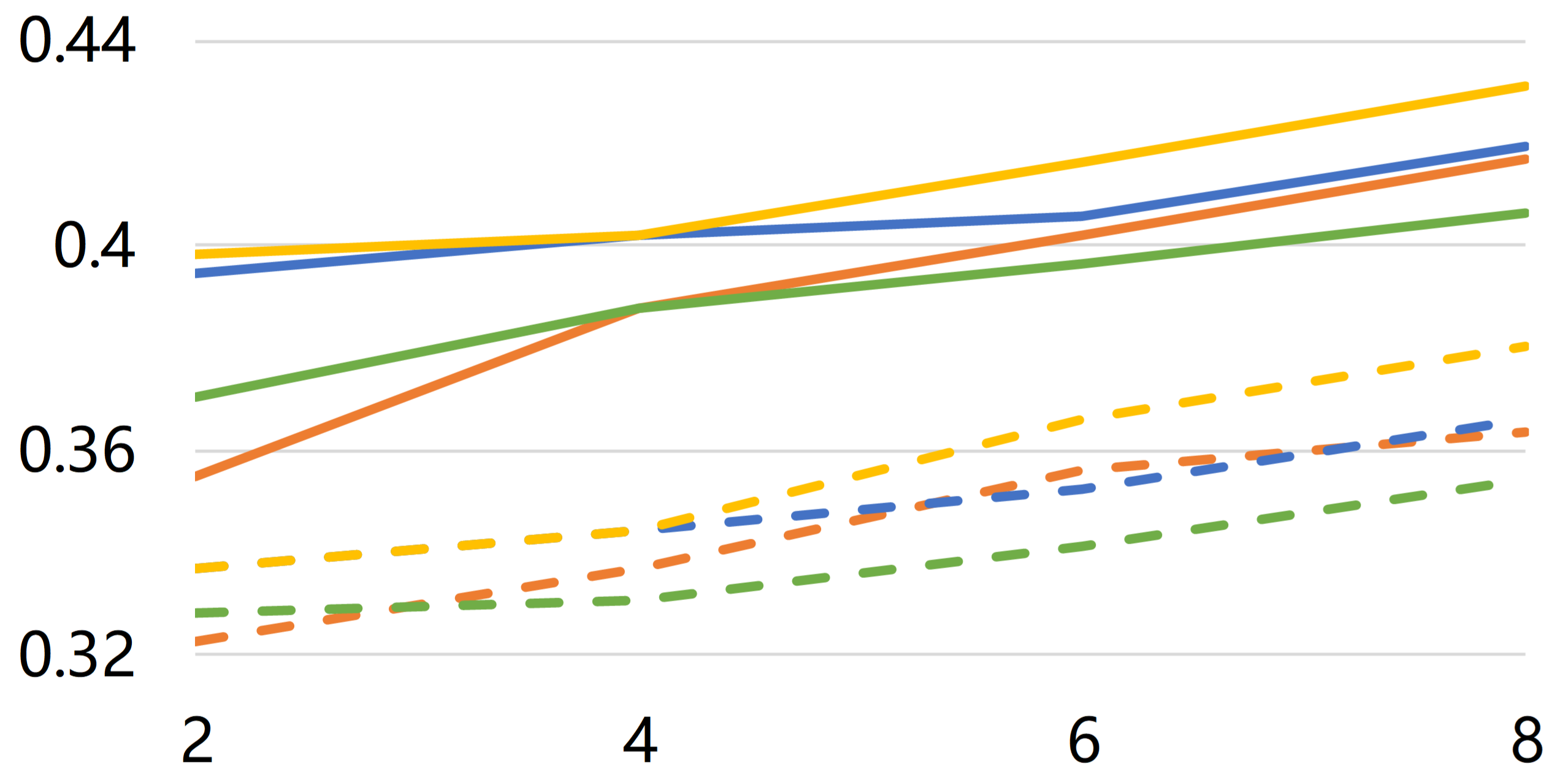}
}
\includegraphics[width=0.85\textwidth]{tuli.png}
\caption{Results of our model and other feature selection methods on bridging tasks with EEG features.}\label{appendix:3}
\end{figure*}

\section{Results of Ablation Study without the BiLSTM Encoder}
The complete results of ablation study are shown in Figure \ref{appendix:1}. From these results, we can find that the variant model without the encoder layer can achieve results similar to the complete framework on both eye-tracking and EEG signals. This indicates that the attention mechanism in our model has good robustness.

\section{Comparison Results of Feature Selection}
The results of different feature selection methods in all bridging tasks are shown in the Figure \ref{appendix:2} and Figure \ref{appendix:3}. Obviously, our method is better than other feature selection method, suggesting that our model can effectively evaluate the degree of importance of features.

\end{document}